\documentclass[journal]{IEEEtran}
\usepackage{amsmath,amssymb,amsfonts}
\usepackage{algorithmic}
\usepackage{graphicx}
\usepackage{textcomp}
\usepackage{xcolor}
\usepackage{url}
\usepackage[numbers]{natbib}
\def\BibTeX{{\rm B\kern-.05em{\sc i\kern-.025em b}\kern-.08em
    T\kern-.1667em\lower.7ex\hbox{E}\kern-.125emX}}

\usepackage{hyperref}       
\usepackage{url}            
\usepackage[utf8]{inputenc} 
\usepackage[T1]{fontenc}    
\usepackage{booktabs}       
\usepackage{amsfonts}       
\usepackage{nicefrac}       
\usepackage{microtype}      
\usepackage{amsthm}
\usepackage{amsmath}
\usepackage{bm}
\usepackage{mathtools}
\usepackage{enumitem}
\usepackage{wrapfig}
\DeclarePairedDelimiterX{\infdivx}[2]{(}{)}{%
  #1\;\delimsize\|\;#2%
}

\newcommand{\Pdata}{\mathbb{P}_{\text{data}}}

\newcommand{\hlam}{\bm{\theta}}
\newcommand{\hlamj}{\theta_j}
\renewcommand{\vec}[1]{\mathbf{#1}}

\begin{document}

\title{Effective Regularization Through\\Evolutionary Loss-Function Metalearning}


\author{
\IEEEauthorblockN{
 Santiago Gonzalez$^1$,\thanks{$^1$Current affiliation: Apple, Inc.}}
\IEEEauthorblockA{\textit{Cognizant AI Lab {\rm and}}
\textit{The University of Texas at Austin}; 
slgonzalez@hey.com\\}
\and
\IEEEauthorblockN{
Xin Qiu,}
\IEEEauthorblockA{\textit{Cognizant AI Lab};
xin.qiu@cognizant.com}\\
\and
\IEEEauthorblockN{
Risto Miikkulainen,}
\IEEEauthorblockA{\textit{Cognizant AI Lab {\rm and}}
\textit{The University of Texas at Austin}; 
risto@cs.utexas.edu}
}

\maketitle 

\begin{abstract}
Evolutionary computation can be used to optimize several different aspects of neural network architectures. For instance, the TaylorGLO method discovers novel, customized loss functions, resulting in improved performance, faster training, and improved data utilization.  A likely reason is that such functions discourage overfitting, leading to effective regularization. This paper demonstrates theoretically that this is indeed the case for TaylorGLO. Learning rule decomposition reveals that evolved loss functions balance two factors: the pull toward zero error, and a push away from it to avoid overfitting. This is a general principle that may be used to understand other regularization techniques as well (as demonstrated in this paper for label smoothing). The theoretical analysis leads to a constraint that can be utilized to find more effective loss functions in practice; the mechanism also results in networks that are more robust (as demonstrated in this paper with adversarial inputs). The analysis in this paper thus constitutes a first step towards understanding regularization, and demonstrates the power of evolutionary neural architecture search in general.
\end{abstract}


\newcommand{\Tx}[1][-\omega_0]{(x_i #1)}
\newcommand{\Ty}[1][-\omega_1]{(y_i #1)}

\section{Introduction}

Regularization is a key concept in deep learning: It guides learning towards configurations that are likely to perform robustly on unseen data. Different regularization approaches originate from intuitive characterization of the learning process and have been shown to be effective empirically \cite{hanson1989comparingbiases,dropout,inceptionv3,real2019regularized}. However, a general theory of the underlying mechanisms, the different types of regularization, and their interactions, still needs to be developed.

This paper takes a first step towards understanding regularization by analyzing one successful such method: evolutionary loss-function optimization. Recently, loss-function optimization has emerged as a new area of neural network metalearning. The general optimization problem is non-differentiable, but well suited for evolutionary approaches: Indeed, new loss functions have already been discovered in this manner, and shown to outperform traditional loss functions \cite{gonzalez2019glo,taylorglo,gonzalez:diss20}. Remarkably, in doing so, evolution discovered a general principle: The evolved loss functions prevent the network from learning predictions with extreme confidence, thus regularizing in a surprising but transparent manner. Moreover, this regularization approach is amenable to theoretical analysis, providing a possible starting point towards understanding regularization more generally.

In order to develop a theory of loss-function regularization, a framework under which the evolved functions can be analyzed and compared, both with each other and with traditional loss functions, is needed. In the framework proposed in this paper, the stochastic gradient descent (SGD) learning rule is decomposed to coefficient expressions that can be symbolically defined for a wide range of loss functions, regardless of their mathematical form. These expressions provide an intuitive understanding of the training dynamics in specific contexts. Within this new framework, the well-known mean squared error (MSE) and cross-entropy loss functions, as well as evolution-discovered Genetic Loss-function Optimization (GLO) / Taylor-expansion GLO (TaylorGLO) loss functions \cite{gonzalez2019glo,taylorglo}, are first analyzed at the null epoch (i.e.\ beginning of learning) and the zero training error regime (i.e.\ end of learning), and the analysis is then generalized for an intermediate point in the training process. This work leads to three main contributions:

\textbf{(1)} A theoretical understanding of how this specific kind of regularization works, i.e.\ how GLO/TaylorGLO avoids becoming overly confident in its predictions. The evolved loss functions balance two factors: the pull toward zero error, and a push away from it to avoid overfitting. This is a general principle that may underlie other regularization techniques as well, as is shown in the paper for label smoothing \cite{inceptionv3}. The results thus suggest that the approach can play a role in developing a more general theory of regularization in the future.

\textbf{(2)} Based on the theory, a practical improvement for loss-function metalearning is identified: an invariant that must hold true for networks to be trainable. This constraint can guide the search process towards good loss functions more efficiently.

\textbf{(3)} Identifying robustness as a new role for regularization; that is, regularization not only improves generalization to unseen examples, but also makes the system more robust against other kinds of variation. This conclusion is drawn experimentally through adversarial samples, and shown to be a likely result of wider decision basins in network performance.

Note that the paper does not aim to compare different regularization methods, nor to demonstrate that loss-function optimization is in some way the best. Instead, it provides an approach to understanding regularization in this special case, i.e. a first step towards developing a more general theory in the future. Indeed, whether other regularization methods address different aspects of performance, synergetically or in a potentially unifiable fashion to loss-function regularization, is a most interesting direction of future research.

Also note that adversarial examples are used to demonstrate that loss-function-based regularization also improves network performance in another dimension: robustness. It is part of developing an understanding of how regularization works. Whether loss-function optimization can be developed further into an actual solution for adversarial robustness is a question for future research.

Thus, the paper demonstrates how regularization arises from loss-function metalearning, and how it can be effective in improving performance. It suggests that a theoretical understanding of regularization is possible, to be generalized to other regularization approaches in the future. In doing so, it demonstrates the power of evolutionary neural architecture search in a general sense, i.e.\ that evolutionary optimization of different aspects of neural network design, beyond simply the network topology, can be highly useful.
General background on regularization approaches and loss-function metalearning is reviewed in Appendix~\ref{ap:background}. The particular method analyzed in this paper, TaylorGLO, is discussed next.

\section{The TaylorGLO method}
\label{sec:taylorglo_method}

\begin{figure}
  \centering
  \includegraphics[width=0.65\linewidth]{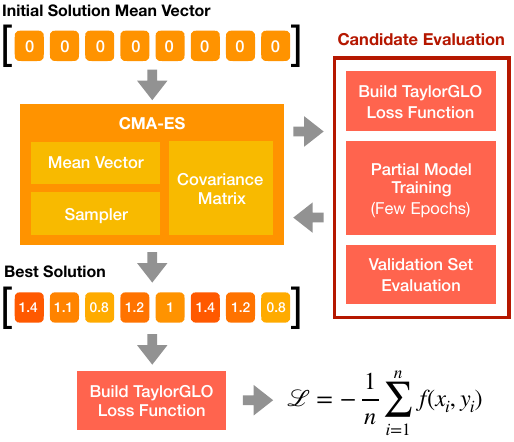}
  \vspace{-0.7em}
  \caption{The TaylorGLO method \cite{taylorglo}. Loss functions are represented by fixed-size vectors whose elements parameterize modified Taylor polynomials. Starting with a population of initially unbiased loss functions (i.e., vectors around the origin), CMA-ES optimizes their Taylor expansion parameters in order to maximize validation accuracy after partial training. The candidate with the highest accuracy is chosen as the final, best solution. This approach biases the search towards functions with useful properties, and is also amenable to theoretical analysis, as shown in this paper.}
  \label{fig:overview}
\end{figure}

TaylorGLO\footnote{Open-source code for TaylorGLO is available at \url{https://github.com/cognizant-ai-labs/taylorglo}.} 
(Figure~\ref{fig:overview}) aims to find the optimal parameters for a loss function represented as a multivariate Taylor expansion. 
The details of this parameterization are described in Appendix~\ref{sec:taylorexpansions}.
The parameters\footnote{The original paper on TaylorGLO \cite{taylorglo} formulated TaylorGLO in terms of $\theta$ rather than $\omega$. This paper uses $\omega$ to avoid overloading notation in later sections.} for a Taylor approximation (i.e., the center point and partial derivatives) are referred to as $\vec{\omega}_{\hat{f}}$: $\vec{\omega}_{\hat{f}} \in \Omega$, $\Omega = \mathbb{R}^{\#_{\text{parameters}}}$. TaylorGLO strives to find the vector $\vec{\omega}_{\hat{f}}^*$
that parameterizes the optimal loss function for a task. Because the values are continuous, as opposed to discrete graphs of the original GLO, it is possible to use continuous optimization methods.

In particular, Covariance Matrix Adaptation Evolutionary Strategy (CMA-ES) \cite{hansen1996cmaes} is a popular population-based, black-box optimization technique for rugged, continuous spaces. CMA-ES functions by maintaining a covariance matrix around a mean point that represents a distribution of solutions. At each generation, CMA-ES adapts the distribution to better fit evaluated objective values from sampled individuals. In this manner, the area in the search space that is being sampled at each step grows, shrinks, and moves dynamically as needed to maximize sampled candidates' fitnesses. TaylorGLO uses the ($\mu/\mu,\lambda$) variant of CMA-ES \cite{hansen2001cmaesmumulambda}, which incorporates weighted rank-$\mu$ updates \cite{hansen2004weightedrankmucmaes} to reduce the number of objective function evaluations needed. 

In order to find $\vec{\omega}_{\hat{f}}^*$, at each generation CMA-ES samples points in $\Omega$. Their fitness is determined by training a model with the corresponding loss function and evaluating the model on a validation dataset. Fitness evaluations may be distributed across multiple machines in parallel and retried a limited number of times upon failure. An initial vector of $\vec{\omega}_{\hat{f}} = \vec{0}$ (that is, parameters that represent a flat function with zero gradients) is chosen as a starting point in the search space to avoid bias. 
For further details on the experimental setup, see Appendix~\ref{sec:experimental_setup_appendix}.



Fully training a model can be prohibitively expensive in many problems. However, performance near the beginning of training is usually correlated with performance at the end of training, and therefore it is enough to train the models only partially to identify the most promising candidates. This type of approximate evaluation is common in metalearning \cite{grefenstette1985genetic, jin2011}. An additional positive effect is that evaluation then favors loss functions that learn more quickly.

For a loss function to be useful, it must have a derivative that depends on the prediction. Therefore, internal terms that do not contribute to $\frac{\partial}{\partial \vec{y}}\mathcal{L}_f(\vec{x},\vec{y})$ can be trimmed away. This step implies that any term $t$ within $f(x_i,y_i)$ with $\frac{\partial}{\partial y_i}t = 0$ can be replaced with $0$.
For example, this refinement simplifies Equation~\ref{eq:k3example}, providing a reduction in the number of parameters from twelve to eight:
\vspace{-0.5em}
\begin{equation}
\label{eq:k3taylorglo}
\begin{aligned}
\mathcal{L}(\vec{x},\vec{y}) = -\frac{1}{n}\sum^n_{i=1} \Big[    \omega_2\Ty + \tfrac{1}{2}\omega_3\Ty^2 \\+ \tfrac{1}{6}\omega_4\Ty^3  + \omega_5\Tx\Ty
 \\+ \tfrac{1}{2}\omega_6\Tx\Ty^2 + \tfrac{1}{2}\omega_7\Tx^2\Ty  \Big] \;.
\end{aligned}
\end{equation}

TaylorGLO has been applied to different benchmark datasets and architectures with standard hyperparameters \cite{taylorglo}. These setups have been heavily engineered and manually tuned by the research community, yet TaylorGLO's evolution was able to discover task-customized loss functions that provide a further statistically significant performance improvement.

Most importantly, TaylorGLO (and GLO earlier) discovered an interesting general principle that transfers across datasets and models. One particular type of loss function, called Baikal for its shape, emerged often in the experiments:
 \vspace{-0.5em}
\begin{equation}
\mathcal{L}_{\text{Baikal}} = -\frac{1}{n}\sum^{n}_{i=0} \log (y_i) - \frac{x_i}{y_i}.
\label{eqn:baikal}
\end{equation}
where $\vec{x}$ represents the ground-truth label, and $\vec{y}$ represents a model's predicted label.\footnote{This is the original GLO notation, which differs slightly from the notation used
in Section~\ref{sec:learning_rule_decomp} and later in this paper.}

Baikal was shown to improve overall classification accuracy, accuracy in low-data settings, and training speed in several settings. It was conjectured to achieve these properties through a form of regularization that ensures that the model does not become overly confident in its predictions. That is, instead of monotonically decreasing the loss when the output gets closer to the correct value, Baikal increases rapidly when the output is almost correct, thus discouraging extreme accuracy.

Building on this foundation, the next section develops a theoretical framework for analyzing Baikal in particular and TaylorGLO loss functions in general.

\section{Characterizing training dynamics}
\label{sec:characterizingtrainingdynamics}

The first step in characterizing training dynamics is to decompose the various learning rules into a canonical form. The decompositions are then first analyzed at the null epoch, i.e.\ the initial state of the learning process, when network weights are similarly distributed. This analysis leads to a constraint on the learning process that can be used to make evolution more effective, as will be discussed in Section~\ref{sec:invariant_section}. Behavior at the opposite end of the training process will then be analyzed, i.e.\ in the zero training error regime. In this regime it is possible to identify optimization biases that lead to implicit regularization. Third, generalizing to the entire training process, a theoretical constraint is derived on the entropy of a network's outputs. This constraint makes it possible to characterize learning in TaylorGLO as an interaction between data fitting and regularization. Fourth, a secondary mechanism for regularization, implicit label smoothing, is identified. TaylorGLO may discover and utilize label smoothing as part of effective loss functions.

\subsection{Learning rule decomposition}

\begin{table}
  \caption{Overview of notation used in this paper.}
  \vspace{-1em}
  \label{tab:notation}
\begin{center}
\footnotesize
\begin{tabular}{cl}
\toprule
\textbf{Symbol} & \textbf{Description}\\
\midrule
	$h(\bm{x}_i,\hlam)$ & The model, with a softmax \\ 
	$h_k(\bm{x}_i,\hlam)$ & The model's $k$th scaled logit \\ 
	$D_{\bm{j}}\left( f \right)$ & The directional derivative of $f$ along $\bm{j}$ \\

    $\Pdata$ & Probability distribution of original data \\
    $\bm{x}_i$ & An input data sample, where $\bm{x}_i\sim \Pdata$ \\
    $\bm{y}_i$ & A label that corresponds to the $\bm{x}_i$ sample \\
    $\eta$ & Learning rate \\
    $n$ & Number of classes \\
    $\hlam$ & A model's trainable parameters \\
    $\bm{\lambda}$ & The loss function's parameters \\    
    $\mathcal{L}(\bm{x}_i,\bm{y}_i,\hlam)$ & The loss function \\
    $\gamma_k(\bm{x}_i,\bm{y}_i,\hlam)$ & Decomposed loss function expression from Equation~\ref{eqn:gamma_decomposition} \\
\bottomrule
\end{tabular}
\end{center}
\vspace{-1em}
\end{table}

By decomposing the learning rule, the contribution of the loss function becomes clear. Comparisons of different loss functions can then be drawn at different stages of the training process. 

An overview of the notation used in this section is given in Table~\ref{tab:notation}. To begin, consider the standard SGD update rule:
\begin{equation}
\hlam \leftarrow \hlam - \eta \nabla_{\hlam} \left( \mathcal{L}(\bm{x}_i,\bm{y}_i,\hlam) \right) .
\end{equation}
where $\eta$ is the learning rate, $\mathcal{L}(\bm{x}_i,\bm{y}_i,\hlam)$ is the loss function applied to the network $h(\bm{x}_i,\hlam)$, $\bm{x}_i$ is an input data sample, $\bm{y}_i$ is the $i$th sample's corresponding label, and $\hlam$ is the set of trainable parameters in the model.
The update for a single weight $\hlamj$ is
\begin{equation}
\label{eqn:sgdsingleweight}
\hlamj \leftarrow \hlamj - \eta D_{\bm{j}} \left( \mathcal{L}(\bm{x}_i,\bm{y}_i,\hlam) \right)      =       \hlamj - \eta \frac{\partial}{\partial s} \mathcal{L}(\bm{x}_i,\bm{y}_i,\hlam + s \bm{j}) \;\bigg\rvert_{s\rightarrow 0},
\end{equation}
where $\bm{j}$ is a basis vector for the $j$th weight. This general learning rule can then be decomposed in a classification context for a variety of loss functions: mean squared error (MSE), the cross-entropy loss function, the general third-order TaylorGLO loss function, and the Baikal loss function. Each decomposition results in a learning rule of the form
\vspace{-0.5em}
\begin{equation}
\label{eqn:gamma_decomposition}
\hlamj \leftarrow \hlamj + \eta \frac{1}{n} \sum^n_{k=1}\left[ \gamma_k(\bm{x}_i,\bm{y}_i,\hlam) D_{\bm{j}} \left( h_k(\bm{x}_i,\hlam) \right)   \right],
\end{equation}
where $\gamma_k(\bm{x}_i,\bm{y}_i,\hlam)$ is an expression that is specific to each loss function.

\newcommand{\dirdivhk}{\frac{\partial}{\partial s} h_k(\bm{x}_i,\hlam + s \bm{j}) }
\newcommand{\hkminuslamone}{\left(h_k(\bm{x}_i,\hlam + s \bm{j}) - \lambda_1 \right) }
\vspace{0.5em}

In Appendix~\ref{sec:learning_rule_decomp}, this decomposition is derived for the four loss functions analyzed in this section: mean squared error (MSE), the cross-entropy loss function, the Baikal loss function, and the general third-order TaylorGLO loss function.

\subsection{Behavior at the null epoch}
\label{sec:nullepoch}
Consider the first epoch of training. Assume all weights are randomly initialized:
\vspace{-0.5em}
\begin{equation}
\forall k \in [1,n], \text{where}\; n\geq 2:\; \mathop{\mathbb{E}}_i \left[ h_k(\bm{x}_i,\hlam) \right] = \frac{1}{n} .
\end{equation}
\vspace{-0.5em}
\\
That is, logits (the neural network's final output activations, which sum to one) are distributed with high entropy. Behavior at the null epoch can then be defined piecewise for target vs.\ non-target logits for each loss function.
\vspace{-0.5em}
\\
~\\
In the case of {\bf Mean squared error (MSE),}
\begin{equation}
\gamma_k(\bm{x}_i,\bm{y}_i,\hlam) = \left\{
        \begin{array}{rl}
           -2n^{-1} & \quad y_{ik} = 0 \\
           2 - 2n^{-1} & \quad y_{ik} = 1 .
        \end{array}
    \right.
\label{eq:mse}
\end{equation}
Since $n\geq2$, the $y_{ik} = 1$ case will always be non-negative, while the $y_{ik} = 0$ case will always be negative. Thus, target scaled logits will be maximized and non-target scaled logits minimized.
\vspace{-0.5em}
\\
~\\
In the case of {\bf Cross-entropy loss,}
\begin{equation}
\gamma_k(\bm{x}_i,\bm{y}_i,\hlam) = \left\{
        \begin{array}{rl}
           0 & \quad y_{ik} = 0 \\
           n & \quad y_{ik} = 1 .
        \end{array}
    \right.
\end{equation}
Target scaled logits are maximized and, consequently, non-target scaled logits minimized as a result of the softmax function.
\vspace{-0.5em}
\\
~\\
Similarly in the case of {\bf Baikal loss,}
\begin{equation}
\gamma_k(\bm{x}_i,\bm{y}_i,\hlam) = \left\{
        \begin{array}{rl}
           n & \quad y_{ik} = 0 \\
           n + n^2 & \quad y_{ik} = 1.
        \end{array}
    \right.
\label{eq:baikal}
\end{equation}
Target scaled logits are maximized and, consequently, non-target scaled logits minimized as a result of the softmax function (since the $y_{ik} = 1$ case dominates).
\vspace{-0.5em}
\\
~\\
In the case of {\bf Third-order TaylorGLO loss,} since behavior is highly dependent on $\bm{\lambda}$, consider the concrete loss function that TaylorGLO discovered for the AllCNN-C model on CIFAR-10 \cite{taylorglo}:
\begin{equation}
\gamma_k(\bm{x}_i,\bm{y}_i,\hlam) = \left\{
        \begin{array}{rl}
           -373.917 - 130.264 \;h_k(\bm{x}_i,\hlam) \\- 11.2188 \;h_k(\bm{x}_i,\hlam)^2 &  y_{ik} = 0 \\
            -372.470735 - 131.47 \;h_k(\bm{x}_i,\hlam) \\- 11.2188 \;h_k(\bm{x}_i,\hlam)^2 &  y_{ik} = 1 .
        \end{array}
    \right.
\end{equation}
Let us substitute $h_k(\bm{x}_i,\hlam) = \frac{1}{n}$ (i.e., the expected value of a logit at the null epoch): 
\begin{equation}
\gamma_k(\bm{x}_i,\bm{y}_i,\hlam) = \left\{
        \begin{array}{rl}
           -373.917 - 130.264 \;n^{-1} \\- 11.2188 \;n^{-2} & \quad y_{ik} = 0 \\
            -372.470735 - 131.47 \;n^{-1} \\- 11.2188 \;n^{-2} & \quad y_{ik} = 1 .
        \end{array}
    \right.
\end{equation}
\vspace{-0.5em}
\\
Since this loss function was found on CIFAR-10, a 10-class image classification task, $n=10$:
\begin{equation}
\gamma_k(\bm{x}_i,\bm{y}_i,\hlam) = \left\{
        \begin{array}{rl}
           -386.9546188 & \quad y_{ik} = 0 \\
            -385.729923 & \quad y_{ik} = 1 .
        \end{array}
    \right.
\end{equation}
Since both cases of $\gamma_k(\bm{x}_i,\bm{y}_i,\hlam)$ are negative, this behavior implies that all scaled logits will be minimized. However, since the scaled logits are the output of a softmax function, and the $y_{ik} = 0$ case is more strongly negative, the non-target scaled logits will be minimized more than the target scaled logits, resulting in a maximization of the target scaled logits.

The desired behavior at the null epoch is clear, and the above evaluated loss functions all exhibit it. However, certain settings for $\bm{\lambda}$ in TaylorGLO loss functions may result in detrimental behavior. Thus, a constraint on $\bm{\lambda}$ can be derived to make sure that it does not happen. Such a constraint is
derived in Appendix~\ref{sec:tayinvariantderivation} and
used to speed up the TaylorGLO search process in Section~\ref{sec:invariant_section}.

Next, the opposite end of the training process is analyzed in order to identify optimization biases with different loss functions. They will lead to understanding of the regularization mechanisms in TaylorGLO, as will be discussed in Section~\ref{sec:regularization}.

\subsection{Biases in the zero training error regime}

Certain biases in optimization imposed by a loss function can be best observed in the case where there is nothing new to learn from the training data. Consider the case where there is zero training error,
that is, $h_k(\bm{x}_i,\hlam) - y_{ik} = 0$. In this case, all $h_k(\bm{x}_i,\hlam)$ can be substituted with $y_{ik}$ in $\gamma_k(\bm{x}_i,\bm{y}_i,\hlam)$, as is done below for the different loss functions. While zero training error is not always possible in practice, this case still approximates what happens the network approaches low training error conditions and provides insight on a loss function's inherent optimization biases.
\vspace{-0.5em}
\\
~\\
\textbf{Mean squared error (MSE):\quad} In this case,
\begin{equation}
    \begin{aligned}
\gamma_k(\bm{x}_i,\bm{y}_i,\hlam) = 2y_{ik} - 2 h_k(\bm{x}_i,\hlam) = 0.
    \end{aligned}
\end{equation}
Thus, there are no changes to the weights of the model once error reaches zero. This observation contrasts with the findings in \cite{blanc2020implicit}, who discovered an implicit regularization effect when training with MSE loss \emph{and} label noise. Notably, this null behavior is representable in a non-degenerate TaylorGLO parameterization, since MSE is itself representable by TaylorGLO with $\bm{\lambda} = [0,0,0,-1,0,2,0,0]$. Thus, this behavior can be leveraged in evolved loss functions.
\vspace{-0.5em}
\\
~\\
\textbf{Cross-entropy loss:\quad}
Since $h_k(\bm{x}_i,\hlam) = 0$ for non-target logits in a zero training error regime, $\gamma_k(\bm{x}_i,\bm{y}_i,\hlam) = \frac{0}{0}$, i.e.\ an indeterminate form. Thus, an arbitrarily-close-to-zero training error regime is analyzed instead, such that  $h_k(\bm{x}_i,\hlam) = \epsilon$ for non-target logits for an arbitrarily small $\epsilon$. Since all scaled logits sum to $1$, $h_k(\bm{x}_i,\hlam) = 1-(n-1)\epsilon$ for the target logit. Let us analyze the learning rule as $\epsilon$ tends towards $0$:
\vspace{-0.5em}
\begin{equation}
\hlamj \leftarrow \hlamj + \lim_{\epsilon\to 0} \eta \frac{1}{n} \sum^n_{k=1} \left\{
        \begin{array}{rl}
            \dfrac{y_{ik}}{\epsilon}  D_{\bm{j}} \left( h_k(\bm{x}_i,\hlam) \right) &  y_{ik} = 0 \\
            \dfrac{y_{ik}}{1-(n-1)\epsilon} D_{\bm{j}} \left( h_k(\bm{x}_i,\hlam) \right)  &  y_{ik} = 1
        \end{array}
    \right.
\end{equation}
\begin{equation}
= \hlamj + \eta \frac{1}{n} \sum^n_{k=1} \left\{
        \begin{array}{rl}
            0 & \quad y_{ik} = 0 \\
            D_{\bm{j}} \left( h_k(\bm{x}_i,\hlam) \right)  & \quad y_{ik} = 1 .
        \end{array}
    \right.
\end{equation}
Intuitively, this learning rule aims to increase the value of the target scaled logits. Since logits are scaled by a softmax function, increasing the value of one logit decreases the values of other logits. Thus, the fixed point of this bias will be to force non-target scaled logits to zero, and target scaled logits to one. In other words, this behavior aims to minimize the divergence between the predicted distribution and the training data's distribution.

\begin{figure*}[!ht]
  \centering
  \vspace*{-2.5ex}
  \begin{minipage}{0.5\textwidth}
  \centering
  \caption{Per-sample attraction towards zero training error with cross-entropy vs.\ TaylorGLO loss functions on CIFAR-10 AllCNN-C models. Each point represents an individual training sample (500 random samples per epoch); its $x$-location indicates the training epoch, and $y$-location the strength with which the loss functions pulls the output towards the correct label, or pushes it away from it. With the cross-entropy loss, these values are always positive, indicating a constant pull towards the correct label for every single training sample. Interestingly, the TaylorGLO values span both the positives and the negatives; at the beginning of training there is a strong pull towards the correct label (dark area on top left), which then changes to more prominent push away from it in later epochs. This plot shows how TaylorGLO regularizes by preventing overconfidence and biasing solutions towards different parts of the weight space.}
  \label{fig:zeroerrforces}
  \end{minipage}
  \hfill
  \begin{minipage}{0.24\textwidth}
  \centering
  \vspace*{1.5ex}
  \includegraphics[width=\textwidth]{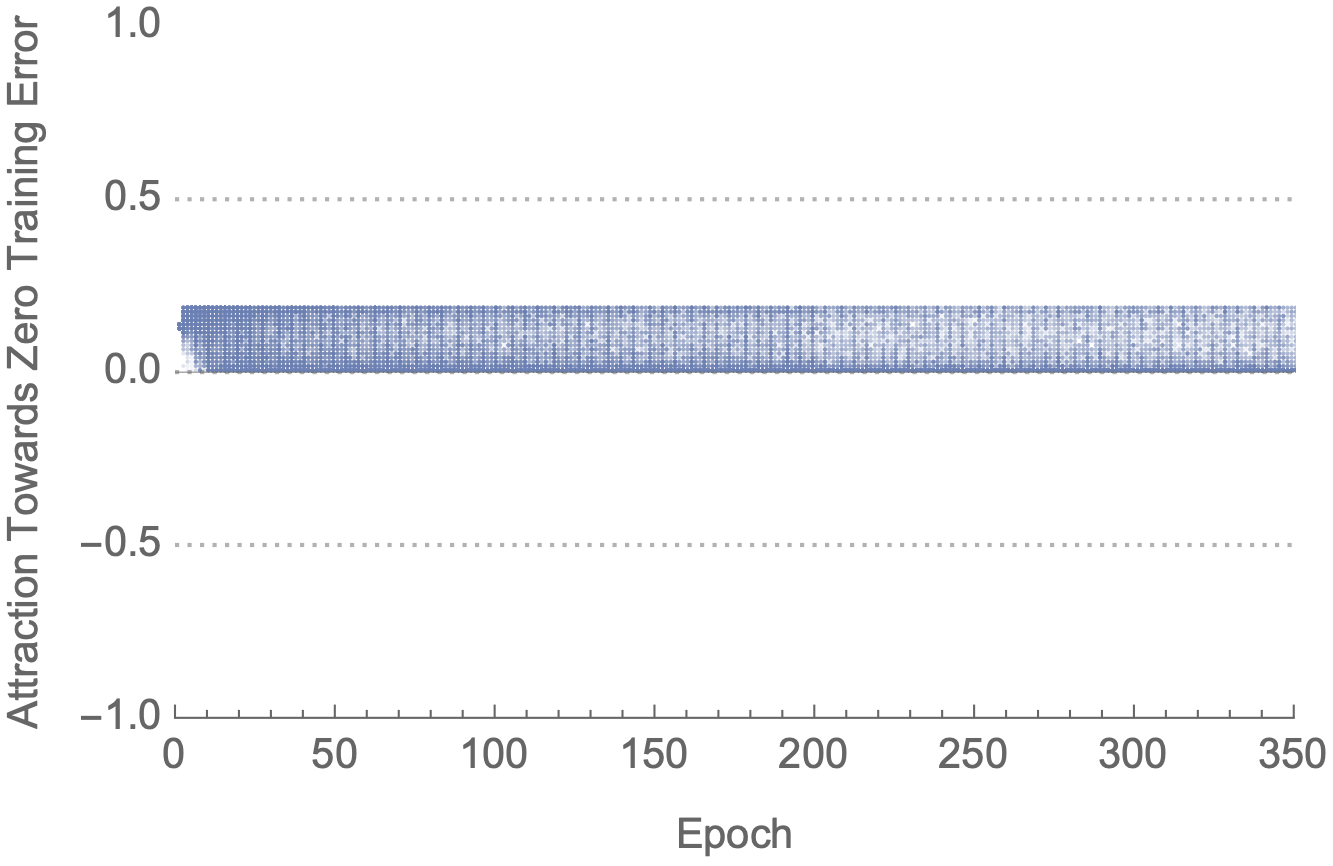}\\
  {\footnotesize $(a)$ Cross-Entropy Loss}
  \end{minipage}
  \hfill
  \begin{minipage}{0.24\textwidth}
  \centering
  \vspace*{1.5ex}
  \includegraphics[width=\textwidth]{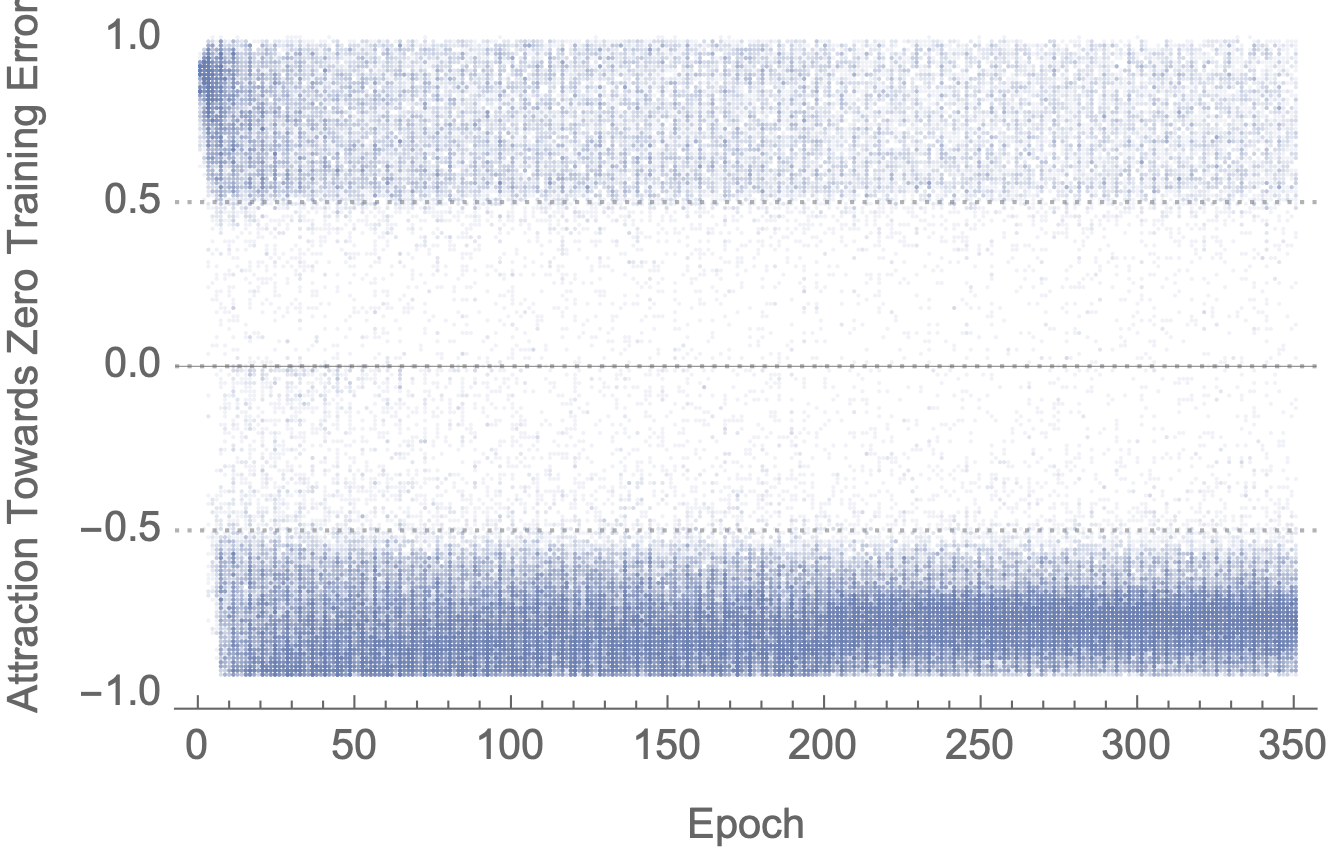}\\
  {\footnotesize $(a)$ TaylorGLO Loss}
  \end{minipage}\\
  \begin{minipage}{1.0\textwidth}
  \end{minipage}
\end{figure*}

TaylorGLO can represent this behavior, and can thus be leveraged in evolved loss functions, through any case where $a=0$ and $b+c>0$. Any $\bm{\lambda}$ where $\lambda_2 = 2 \lambda_1 \lambda_3 + \lambda_5 \lambda_0 - 2\lambda_1\lambda_6\lambda_0 - \lambda_7 \lambda_0^2 - 3 \lambda_4 \lambda_1^2$ represents such a satisfying family of cases. Additionally, TaylorGLO allows for the strength of this bias to be tuned independently from $\eta$ by adjusting the magnitude of $b+c$.
\vspace{-0.5em}
\\
~\\
\textbf{Baikal loss:\quad}
The Baikal loss function results in infinite gradients at zero training error, rendering it unstable, even if using it to fine-tune from a previously trained network that already reached zero training error. However, the zero-error regime is irrelevant with Baikal because it cannot be reached in practice:
\vspace{-0.5em}
\\
~\\
\emph{\textbf{Theorem~1.} Zero training error regions of the weight space are not attractors for the Baikal loss function.}
\vspace{-0.5em}
\\
~\\
The reason is that if a network reaches reaches a training error that is arbitrarily close to zero, there is a repulsive effect that biases the model's weights away from zero training error. Proof of this theorem is in Appendix~\ref{sec:baikalzeroerrnotattractor}.
\vspace{-0.5em}
\\
~\\
\textbf{Third-order TaylorGLO loss:\footnote{Note that in the basic classification case, $\forall w\in \mathbb{N}_1: y_{ik} = y_{ik}^w$, since $y_{ik} \in \{0,1 \}$; This provides an intuition for why higher-order TaylorGLO loss functions do not provide fundamentally different behavior, beyond a more overparameterized search space, and thus no improvements in performance, over third-order loss functions.}\quad}
According to Equation~\ref{eqn:tayk3_gamma_ccomb}, in the zero-error regime,
$\gamma_k(\bm{x}_i,\bm{y}_i,\hlam)$ can be written as a linear combination  $\gamma_k(\bm{x}_i,\bm{y}_i,\hlam) = a + b y_{ik} + c y_{ik}^2$,
where
\begin{eqnarray}
a &=& \lambda_2 - 2 \lambda_1 \lambda_3 - \lambda_5 \lambda_0 + 2\lambda_1\lambda_6\lambda_0 + \lambda_7 \lambda_0^2 + 3 \lambda_4 \lambda_1^2 \\
b &=& 2\lambda_3 - 2\lambda_6 \lambda_0 - 2\lambda_1\lambda_6 + \lambda_5 - 2\lambda_7\lambda_0 - 6\lambda_4\lambda_1 \\
c &=& 2\lambda_6 + \lambda_7 + 3\lambda_4 .
\end{eqnarray}
The learning rule thus becomes
\vspace{-0.5em}
\begin{equation}
\label{eqn:tayk3_zeroerrorclassrule}
\hlamj \leftarrow \hlamj + \eta \frac{1}{n} \sum^n_{k=1} \left\{
        \begin{array}{rl}
            a D_{\bm{j}} \left( h_k(\bm{x}_i,\hlam) \right) & \quad y_{ik} = 0 \\
            (a+b+c) D_{\bm{j}} \left( h_k(\bm{x}_i,\hlam) \right)  & \quad y_{ik} = 1 .
        \end{array}
    \right.
\end{equation}

As a concrete example, consider again the TaylorGLO loss function for AllCNN-C on CIFAR-10. It had $a=-373.917$, $b=-129.928$, $c=-11.3145$. Notably, all three coefficients are negative, i.e.\ all changes to $\theta_j$ are a negatively scaled values of $D_{\bm{j}} \left( h_k(\bm{x}_i,\hlam) \right)$, as can be seen from Equation~\ref{eqn:tayk3_zeroerrorclassrule}.  Thus, there are two competing processes in this learning rule: one that aims to minimize all non-target scaled logits (decreasing the scaled logit distribution's entropy), and one that aims to minimize the target scaled logit (increasing the scaled logit distribution's entropy). The processes conflict with each other since logits are scaled through a softmax function. These processes can shift weights in a particular way while maintaining zero training error, which results in implicit regularization. If, however, such shifts in this zero training error regime do lead to misclassifications on the training data, $h_k(\bm{x}_i,\hlam)$ would no longer equal $y_{ik}$, and a non-zero error regime's learning rule would come into effect. It would strive to get back to zero training error with a different $\hlam$.

Similarly to Baikal loss, a training error of exactly zero is not an attractor for some third-order TaylorGLO loss functions
(this property can be seen through an analysis similar to that in Appendix~\ref{sec:baikalzeroerrnotattractor}).
The zero-error case would occur in practice only if this loss function were to be used to fine tune a network that truly has a zero training error. It is, however, a useful step in characterizing the regularization in TaylorGLO, as will be seen in the next section.

\subsection{Data fitting vs.\ regularization throughout learning}
\label{sec:regularization}

In order to characterize regularization throughout the training process, we need to understand how specific training samples affect a network's trainable parameters. Under what gradient conditions does a network's softmax function transition from increasing the entropy in the output distribution (i.e.\ regularization) to decreasing it (i.e.\ fitting to the data)? Let us analyze the case where all non-target logits have the same value, $\frac{\epsilon}{n-1}$, and the target logit has the value $1-\epsilon$ (i.e. all non-target classes have equal probabilities).
\vspace{-0.5em}
\\
~\\
\label{thm:softmaxentropy}
\emph{\textbf{Theorem~2.}
The change in entropy is proportional to}
\begin{equation}
\label{eqn:thmentropyreductionstrength}
	\dfrac{ \epsilon (\epsilon - 1) \left( \mathrm{e}^{\epsilon (\epsilon - 1) (\gamma_{\neg T} - \gamma_T)}      -       \mathrm{e}^{\frac{\epsilon (\epsilon - 1) \gamma_T (n-1) + \epsilon \gamma_{\neg T} (\epsilon (n-3) + n - 1)}{(n-1)^2}}               \right)       }{   (\epsilon - 1)\; \mathrm{e}^{\epsilon (\epsilon - 1) (\gamma_{\neg T} - \gamma_T)} - \epsilon\; \mathrm{e}^{\frac{\epsilon (\epsilon - 1) \gamma_T (n-1) + \epsilon \gamma_{\neg T} (\epsilon (n-3) + n - 1)}{(n-1)^2}}            }
\end{equation}
where $\gamma_{\neg T}$ is the value of $\gamma_j$ for non-target logits, and $\gamma_T$ for the target logit.
\vspace{-0.5em}
\\
~\\
Thus, values less than zero imply that entropy is increased, values greater than zero imply that it is decreased, and values equal to zero imply that there is no change. The proof of this theorem is in
Appendix~\ref{sec:softmaxentropyderivation}.



The size of reduction in entropy in Theorem~2 can also be thought of as a measure of the strength of the attraction towards zero training error regions of the parameter space (i.e., shrinking non-target logits and growing target logits imply reduced training error). This strength can be calculated for individual training samples during any part of the training process, leading to the insight that the process results from competing ``push'' and ``pull'' forces. This theoretical insight, combined with empirical data from actual training sessions, explains how different loss functions balance data fitting and regularization.

Figure~\ref{fig:zeroerrforces} provides one such example on AllCNN-C models \cite{allcnn} trained on CIFAR-10 \cite{krizhevsky2009learning} with cross-entropy vs.\ custom TaylorGLO loss functions. Scaled target and non-target logit values were logged for every sample at every epoch and used to calculate respective $\gamma_T$ and $\gamma_{\neg T}$ values. These values were then substituted into Equation~\ref{eqn:thmentropyreductionstrength} to get the strength of bias towards zero training error. 

The cross-entropy loss exhibits a tendency towards zero training error for every single sample, as expected. The TaylorGLO loss, however, has a much different behavior: initially, there is a much stronger pull towards zero training error for all samples---which leads to better generalization \cite{yao2007early,li2019biglr}---after which a stratification occurs, where the majority of samples are repelled, and thus biased towards a different region of the weight space with better performance characteristics. 


\begin{figure*}
  \centering
  \begin{minipage}{0.24\textwidth}
  \centering
  \includegraphics[width=\textwidth]{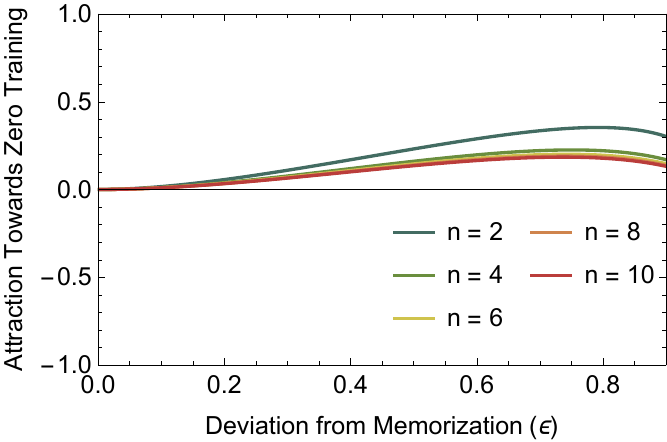}\\
  {\footnotesize $(a)$ Cross-Entropy Loss}
  \end{minipage}
  \hfill
  \begin{minipage}{0.24\textwidth}
  \centering
  \includegraphics[width=\textwidth]{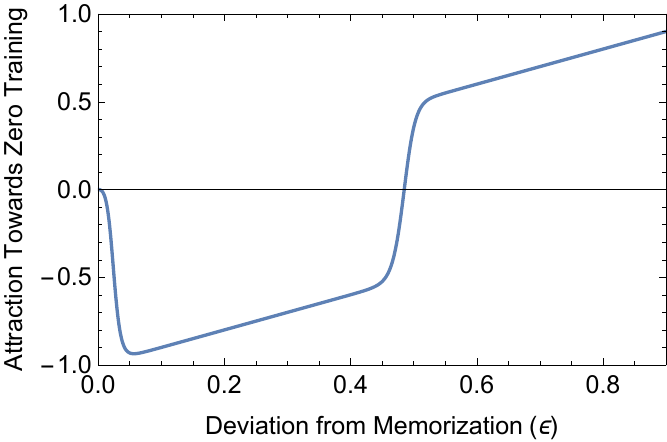}\\
  {\footnotesize $(b)$ TaylorGLO Loss}
  \end{minipage}
  \hfill
  \begin{minipage}{0.24\textwidth}
  \centering
  \includegraphics[width=\textwidth]{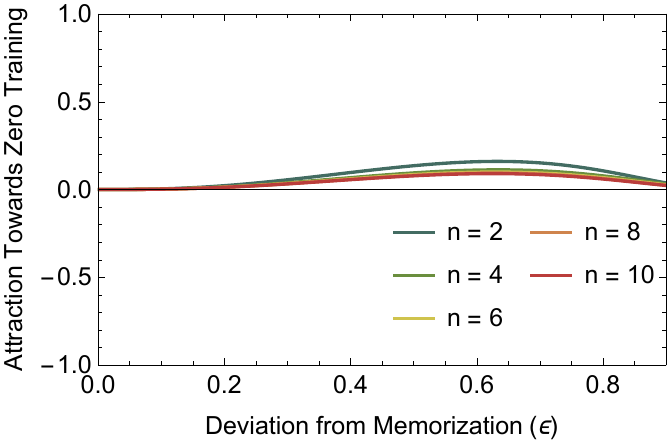}\\
  {\footnotesize $(c)$ MSE Loss}
  \end{minipage}
  \hfill
  \begin{minipage}{0.24\textwidth}
  \centering
  \includegraphics[width=\textwidth]{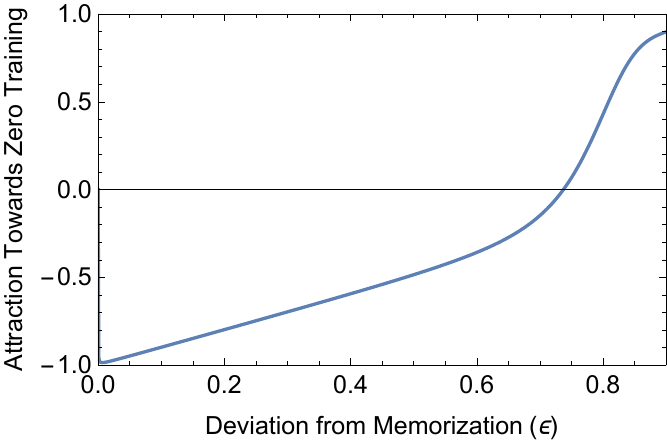}\\
  {\footnotesize $(d)$ Baikal Loss}
  \end{minipage}
  \caption{Attraction towards zero training error with different loss functions. Each loss function has a characteristic curve---plotted using Equation~\ref{eqn:thmentropyreductionstrength}---that describes zero training error attraction dynamics for individual samples given their current deviation from perfect memorization, $\epsilon$. Plots (a) and (b) only have the $n=10$ case plotted, i.e. the 10-class classification case for which they were evolved. Cross-entropy (a) and MSE (c) loss functions have positive attraction for all values of $\epsilon$. In contrast, the TaylorGLO loss function for CIFAR-10 on AllCNN-C (b) and the Baikal loss function (d) both have very strong attraction for weakly learned samples (on the right side), and repulsion for highly confidently learned samples (on the left side). Thus, this illustration provides a graphical intuition for the regularization that TaylorGLO and Baikal loss functions establish.}
  \label{fig:entropy_constraint_plots}
  \vspace{-1em}
\end{figure*}

The strength of the attraction towards zero training error regions of the parameter space (described in Theorem~2) can be plotted---for any given number of classes $n$---at different $\epsilon$ values using the $\gamma_T$ and $\gamma_{\neg T}$ values from a particular loss function. These characteristic curves for four specific loss functions are shown in Figure~\ref{fig:entropy_constraint_plots}.

Both Baikal and TaylorGLO loss functions have significantly different attraction curves than the cross-entropy and mean squared error loss functions. Cross-entropy and mean squared error always exhibit positive attraction to zero training error. Conversely, TaylorGLO and Baikal exhibit this positive attraction behavior only for samples that are weakly memorized; well memorized samples produce a repulsive effect instead. This difference is what contributes to both metalearned loss functions' regularizing effects, where overconfidence is avoided.

The push-pull principle is the core of the regularization theory emerging from the analysis of evolved loss functions. It is a general principle though, and may serve as a foundation for developing a general theory of regularization in the future. As a first step, it is shown to apply to a traditional method of label smoothing.

\subsection{Regularization through implicit label smoothing}

In the previous section, TaylorGLO loss functions were shown to provide regularization through dynamic biases that are imparted throughout the training process. 
This section shows how TaylorGLO can implicitly represent label smoothing \cite{inceptionv3}, suggesting that it may be based on similar principles.

Consider a setup with standard label smoothing, controlled by hyperparameter $\alpha \in (0,1)$, such that the target value in any $\bm{y}_i$ is $1-\alpha\frac{n-1}{n}$ rather than $1$, and non-target values are $\frac{\alpha}{n}$ rather than $0$.
\vspace{-0.5em}
\\
~\\
\emph{\textbf{Theorem~3.} 
For any $\bm{\lambda}$ and any $\alpha \in (0,1)$, there exists a $\bm{\hat{\lambda}}$ such that the behavior imposed by $\bm{\hat{\lambda}}$ without explicit label smoothing is identical to the behavior imposed by $\bm{\lambda}$ \emph{with} explicit label smoothing.
}
~\\
That is, any degree of label smoothing can be implicitly represented for any TaylorGLO loss function.
Thus, the analysis extends to label smoothing as well and may explain certain aspects of TaylorGLO loss functions' regularization. In a similar manner, it may be possible to analyze other regularization methods in the future, eventually leading to a general theory of regularization.
The proof of this theorem is in
Appendix~\ref{sec:labelsmoothinggeneraltaylor}.
\vspace{-0.5em}
\\
~\\
Even though the main goal of the theoretical analysis was to understand the regularization mechanisms in loss-function metalearning, it also leads to a surprising insight that allows improving the search for useful loss functions in practice, as will be discussed next.

\section{Invariant on TaylorGLO parameters}
\label{sec:invariant_section}

As mentioned in Section~\ref{sec:nullepoch}, there are many different instances of $\bm{\lambda}$ for which models are untrainable. One such case, albeit a degenerate one, is $\bm{\lambda} = \bm{0}$ (i.e., a function with zero gradients everywhere).  Given the training dynamics at the null epoch (characterized in Section~\ref{sec:nullepoch}), more general constraints on $\bm{\lambda}$ can be derived, resulting in the following theorem:
\vspace{-0.5em}
\\
~\\
\emph{\textbf{Theorem~4.}
A third-order TaylorGLO loss function is not trainable if the following constraints on $\bm{\lambda}$ are satisfied:}
\vspace{-0.4em}
\begin{eqnarray}
c_1 + c_y + c_{yy} + \dfrac{c_h+c_{hy}}{n} + \dfrac{c_{hh}}{n^2}  < \\ \left(n-1\right)\left(c_1 + \dfrac{c_h}{n} + \dfrac{c_{hh}}{n^2} \right) \\
c_y + c_{yy} + \dfrac{c_{hy}}{n}  <\\ \left(n-2\right)\left(c_1 + \dfrac{c_h}{n} + \dfrac{c_{hh}}{n^2} \right).
\end{eqnarray}
The proof of this theorem is in Appendix~\ref{sec:tayinvariantderivation}.
These constraints are useful because their inverse can be used as an invariant during loss function evolution. That is, they can be used to identify entire families of loss function parameters that do not result in a viable loss function, rule them out during search, and thereby make the search more effective. More specifically, before each candidate $\bm{\lambda}$ is evaluated, it is checked for conformance to the invariant. If the invariant is violated, the algorithm can skip that candidate's validation training and simply assign a fitness of zero. However, due to the added complexity that the invariant imposes on the fitness landscape, a larger population size is needed for evolution within TaylorGLO to be more stable. Practically, a doubling of the population size from 20 to 40 works well.

\begin{figure*}
  \centering
  \begin{minipage}{0.5\textwidth}
  \vspace*{-0ex}
  \centering
  \caption{Test-set accuracy of loss functions discovered by TaylorGLO with and without an invariant constraint on $\bm{\lambda}$. Models were trained on the loss function that had the highest validation accuracy during the TaylorGLO evolution. All averages are from ten separately trained models and $p$-values are from one-tailed Welch's $t$-Tests. Standard deviations are shown in parentheses. The invariant allows focusing metalearning to viable areas of the search space, resulting in better loss functions.}
  \label{tab:results}
  \end{minipage}
  \hfill
  \begin{minipage}{0.48\textwidth}
  \vspace*{1.5ex}
  \footnotesize
  \centering
  \begin{tabular}{l@{~}c@{~}c@{~}c}
    \toprule
    Task and Model&TaylorGLO Acc.&\textbf{+ Invariant}&$p$-value\\
    \midrule
		CIFAR-10, AlexNet \footnotemark[1] 										   & 0.7901 (0.0026) & \textbf{0.7933 (0.0026)} & 0.0092\\
    CIFAR-10, PreResNet-20 \footnotemark[2] 									   & 0.9169 (0.0014) & 0.9164 (0.0019) & 0.2827\\
    CIFAR-10, AllCNN-C \footnotemark[3] 										   & 0.9271 (0.0013) & \textbf{0.9290 (0.0014)} & 0.0004\\
 \bottomrule
\end{tabular}
{\scriptsize
\\
\footnotemark[1] \cite{NIPS2012_4824} \;
\footnotemark[2] \cite{preresnet} \;
\footnotemark[3] \cite{allcnn} 
}
\end{minipage}\\
  \begin{minipage}{1.0\textwidth}
\vspace*{-3ex}
{\bf }
\end{minipage}
\end{figure*}

\begin{figure*}
  \centering
  
  \begin{minipage}{0.24\textwidth}
  \centering
  \includegraphics[width=0.9\textwidth]{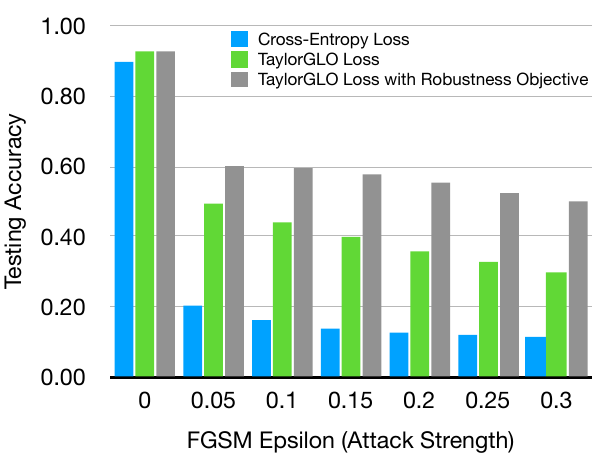}\\
  {\footnotesize $(a)$ AllCNN-C}
  \end{minipage}
  \hfill
  \begin{minipage}{0.24\textwidth}
  \centering
  \includegraphics[width=0.9\textwidth]{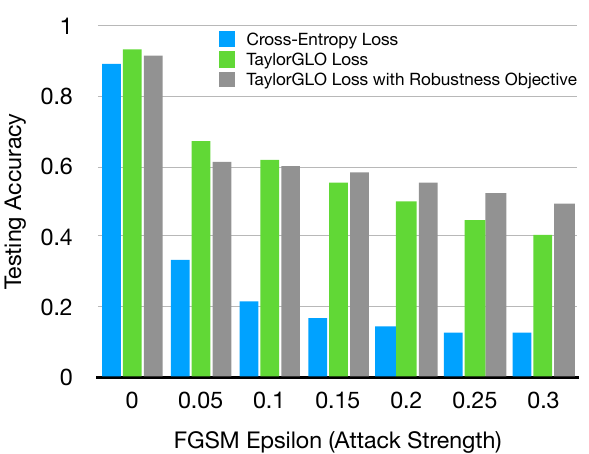}\\
  {\footnotesize $(b)$ AllCNN-C with Cutout}
  \end{minipage}
  \hfill
  \begin{minipage}{0.24\textwidth}
  \centering
  \includegraphics[width=0.9\textwidth]{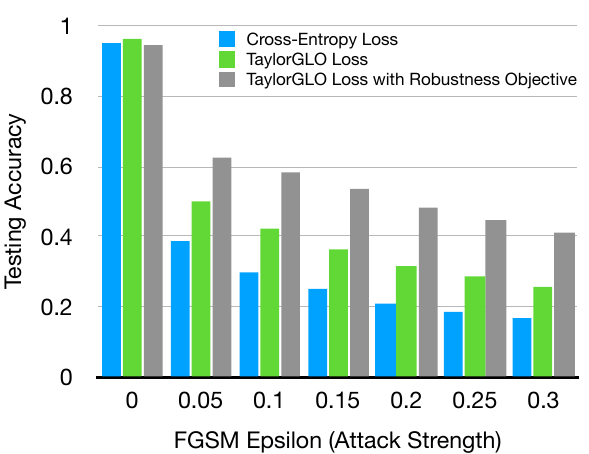}\\
  {\footnotesize $(c)$ Wide ResNet 16-8}
  \end{minipage}
  \hfill
  \begin{minipage}{0.24\textwidth}
  \centering
  \includegraphics[width=0.9\textwidth]{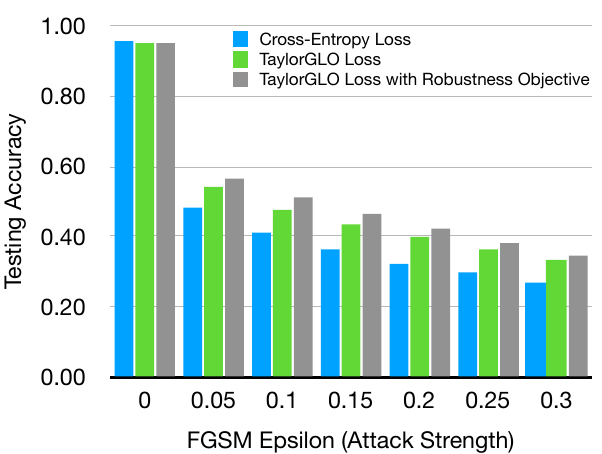}\\
  {\footnotesize $(d)$ Wide ResNet 28-5}
  \end{minipage}
  \vspace{-0.5ex}
  \caption{Robustness of TaylorGLO loss functions against FGSM adversarial attacks with various architectures on CIFAR-10. For each architecture, the blue bars represent accuracy achieved through training with the cross-entropy loss, green bars with a TaylorGLO loss, and gray bars with a TaylorGLO loss specifically evolved in the adversarial attack environment. The leftmost points on each plot represent evaluations without adversarial attacks. TaylorGLO regularization makes the networks more robust against adversarial attacks, and this property can be further enhanced by making it an explicit goal in evolution.}
  \label{fig:adversarial}
\end{figure*}

Figure~\ref{tab:results} presents results from TaylorGLO runs with and without the invariant on the CIFAR-10 image classification benchmark dataset \cite{krizhevsky2009learning} with various architectures. 
Standard training hyperparameters from the references were used for each architecture. Notably, the invariant allows TaylorGLO to discover loss functions that have statistically significantly better performance in many cases and never a detrimental effect, aside from a larger population size, and thus total computational cost. These results demonstrate that the theoretical invariant is useful in practice, and should become a standard in TaylorGLO applications. 

\section{Robustness with Regularization}
\label{sec:adversarial_robustness}

TaylorGLO loss functions discourage overconfidence, i.e.\ their activations are less extreme and vary more smoothly with input. As a result, the networks are likely to generalize better to unseen inputs, but there is also another potential advantage: Such encodings may be more robust against noise, damage, and other imperfections in the data and in the network execution. This hypothesis will be tested experimentally in this section using adversarial inputs, and an empirical explanation will be given in terms of the flatness of loss-surface minima.

\subsection{Evaluation with adversarial inputs}

Adversarial attacks elicit incorrect predictions from a trained model by changing input samples in small ways that can even be imperceptible. They are generally classified as ``white-box'' or ``black-box'' attacks, depending on whether the attacker has access to the underlying model or not, respectively. Naturally, white-box attacks are more powerful at overwhelming a model. One such white-box attack is the Fixed Gradient Sign Method (FGSM) \cite{goodfellow2015explaining}: following evaluation of a dataset, input gradients are taken from the network following a backward pass. Each individual gradient has its sign calculated and scaled by an $\epsilon$ scaling factor that determines the attack strength. These values are added to future network inputs with an $\epsilon$ scaling factor, causing misclassifications.

Figure~\ref{fig:adversarial} shows how robust networks with different loss functions are against FGSM attacks of various strengths. In the first experiment, AllCNN-C and Wide ResNet 28-5 \cite{wideresnet} networks were trained on CIFAR-10 with TaylorGLO and cross-entropy loss; indeed TaylorGLO outperforms the cross-entropy loss models significantly at all attack strengths. Note that in this case, loss functions were evolved simply to perform well, and robustness against adversarial inputs emerged as a side benefit.

An interesting further question emerges: Could adversarial inputs be used to guide loss-function evolution to increase robustness further? Since TaylorGLO uses non-differentiable metrics as objectives in its search process, the traditional validation accuracy objective can be replaced with validation accuracy at a particular FGSM attack strength. This approach was taken in the second experiment, also shown in Figure~\ref{fig:adversarial}. Remarkably, loss functions found in this manner outperform both the previous TaylorGLO loss functions and the cross-entropy loss. 

Thus, these results suggest that TaylorGLO regularization leads to a robust encoding, and such robustness can be further improved by making it an explicit goal in loss-function optimization. Since TaylorGLO does not require differentiable objectives, any measure of robustness (e.g., model accuracy under an adversarial attack) can be targeted for optimization. Where the robustness is coming from will be demonstrated next.


\subsection{Foundation of robustness}
\label{sec:basins}

TaylorGLO loss functions were previously observed to result in trained networks with flatter, lower minima in the weight space \cite{taylorglo}. This provides an intuitive explanation for the robustness: Their performance is less sensitive to small perturbations in the learned parameters.
Since TaylorGLO loss functions that were discovered against an adversarial objective were even more robust, what do their minima look like?

Model performance can be plotted along a random slice $[-1,1]$ of the weight space using a loss surface visualization technique \cite{losslandscape}. The random slice vector is normalized in a filter-wise manner to accommodate network weights' scale invariance, thus ensuring that visualizations for two separate models can be compared. As a result of the randomness, this slice is unbiased and should take all parameters into account, to a degree. It can therefore be used to perturb trainable parameters systematically.

\begin{figure}
  \centering
  \begin{minipage}{0.35\columnwidth}
  \vspace*{-3ex}
  \caption{Comparing accuracy basins of AllCNN-C with cross-entropy, TaylorGLO, and adversarially robust TaylorGLO loss functions on CIFAR-10. Basins are plotted along only one perturbation direction for clarity, using the loss surface visualization technique~~of~~\cite{losslandscape}.~~While}
  \label{fig:adversarial_accuracy_basins}
  \end{minipage}
  \hfill
  \begin{minipage}{0.63\columnwidth}
  \vspace*{-1ex}
  \includegraphics[width=\textwidth]{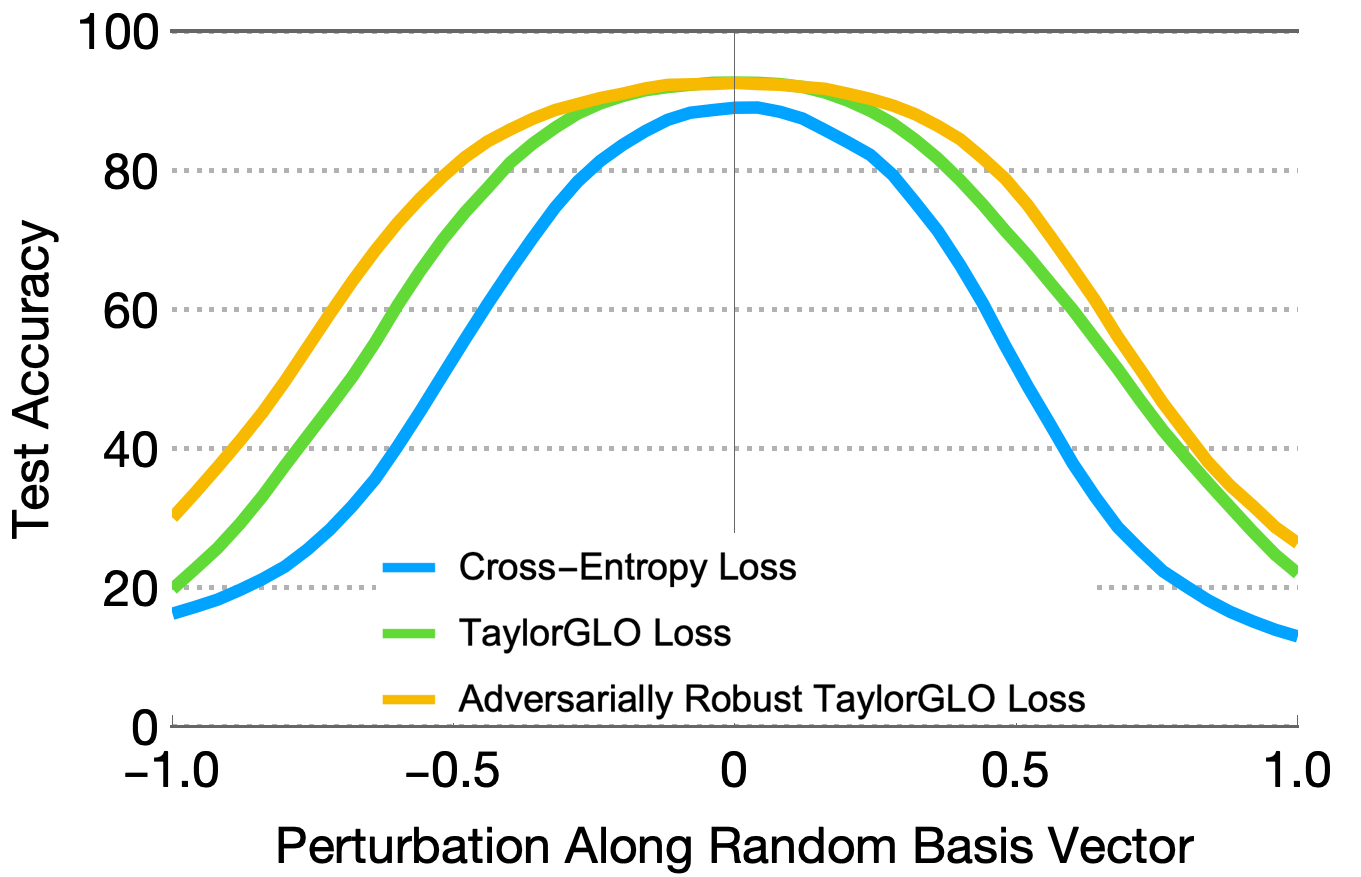}\\
  \vspace{-0.5em}
\end{minipage}
\renewcommand{\baselinestretch}{0.25}
  \begin{minipage}{1.0\columnwidth}
{\footnotesize the adversarially robust TaylorGLO loss function leads to the same accuracy\\[-0.75ex]as the standard one,~~it has a wider, flatter minima.~~~This result suggests that\\[-0.75ex]the TaylorGLO loss function evolved to be robust against adversarial attacks\\[-0.75ex]is more robust in general, even when adversarial attacks are of no concern.}
  \vspace{-1em}
\end{minipage}\\
\end{figure}

When AllCNN-C is trained with an adversarially robust versus a standard TaylorGLO loss function, its absolute accuracy is the same. However, the minimum is wider and flatter (Figure~\ref{fig:adversarial_accuracy_basins}). This result suggests that it may be advantageous to evaluate TaylorGLO against an adversarial performance metric, even when the target application does not include adversarial attacks.


\section{Discussion and Future Work}

This paper builds on an intriguing earlier empirical result: When set to improve a performance objective, evolution of activation functions discovered a regularization mechanism that is surprising yet powerful and compelling. While regularization in general is a complex and poorly understood subject, this mechanism allows gaining insight into it through theoretical analysis. First, the paper developed a learning-rule decomposition framework that makes it possible to characterize and compare the behavior of various loss functions on full-size models. Second, it demonstrated that the mechanism results from balancing the pull towards zero error and the push away from overfitting. Third, it showed that the mechanism may account for a previously-known regularization technique of label smoothing. Fourth, the analysis led to discovering an invariant that can be used in practice to constrain search for better loss functions in the future. Fifth, it showed that this regularization leads to robust encodings, which can be further enhanced through adversarial training. Thus, the study improves our understanding of regularization and the power of evolution to discover it, and provides a foundation for characterizing, comparing, and discovering improved loss functions in the future.

More generally, loss function optimization is one technique in the general metalearning toolbox. It is orthogonal to other techniques, such as neural architecture search, hyperparameter optimization, and activation function metalearning. An interesting direction for future work would be to develop methods that utilize multiple such techniques together, finding synergies between them. For instance, it is likely that certain loss functions work best with certain activation functions. Similarly, loss functions are optimized for a given architecture, and it is possible that those architectures could be optimized to take better advantage of the loss functions. These optimizations could even take place dynamically, while the system is being trained, taking advantage of the different dynamics at different stages of learning \cite{jaderberg2017population,liang:gecco21,bingham:nn22}.

Even though loss-function metalearning is most powerful when it takes advantage of the specific architecture and task, it is also possible to discover general loss functions that work well in several settings. The Baikal loss is already an example of such a general class of functions. In future evolutionary experiments, loss functions can be evaluated across multiple settings, thus rewarding generality. It may also be possible to discover categories of functions that work in different settings, such as wide vs.\ deep, simple vs.\ complex, CNN vs.\ transformer architectures, or noisy vs.\ noiseless, classification vs.\ prediction, sequential vs.\ mapping domains. It may be possible to identify a collection of fundamental loss functions from which an appropriate one can be chosen for each setting, and then evolve it further to obtain superior customized performance.

While the adversarial experiments were used to demonstrate the general robustness of encodings caused by evolved loss functions, it may be possible to develop this approach further towards a general shield against adversarial attacks. Loss functions could be evolved against a variety of attacks, and properties of different attack-specific, evolved loss functions compared. Evolution may be able to discover principles of attack resistance this way, similarly to discovering principles of regularization in the current paper. Such loss functions that harden the machine learning system against adversarial attacks could be combined with other state-of-the-art approaches to build a general shield. This is an interesting opportunity for further research.

\section{Conclusion}

Regularization has long been an important, albeit poorly understood, aspect of training deep neural networks. This paper contributed a theoretical and empirical understanding of one recent and compelling family of regularization techniques: loss-function metalearning. A theoretical framework for representing different loss functions was first developed in order to analyze their training dynamics in various contexts. This framework was applied to TaylorGLO loss functions that were discovered by evolution, demonstrating that they implement a push-pull mechanism that serves as an evolved guard against overfitting. Remarkably, evolution discovered this principle not as a goal of its own, but simply in order to improve performance. The principle was shown to relate to the previous regularization technique of label smoothing, suggesting that in the future it may serve as a stepping stone for developing a general theory of regularization. Two practical opportunities emerged from this analysis: (1) filtering based on an invariant was shown to improve the search process, and (2) training with adversarial inputs was shown to amplify the robustness of the regularized encodings, improving performance overall. The results thus provide theoretical and practical insight into regularization and loss-function metalearning, and demonstrate the power of evolution in scientific discovery and neural architecture search.

\section*{Note}
A shorter version of this paper appeared in the Proceedings of the IEEE Congress on Evolutionary Computation (2025). This paper includes appendices, expanded references, and corrections in Equations~\ref{eq:mse} and~\ref{eq:k3example} and in their descriptions, in the explanations of Equations~\ref{eq:baikal} and~\ref{eqn:tayk3_zeroerrorclassrule}, and in the first paragraph of Section~\ref{sec:regularization}.


\bibliographystyle{ieeetran}
\bibliography{prop}

\onecolumn
\appendices

\section{Background}
\label{ap:background}

Regularization traditionally refers to methods for encouraging smoother mappings from model inputs to outputs by adding a regularizing term to the objective function, i.e., to the loss function in neural networks.  It can be defined more broadly, however, e.g.\ as ``any modification we make to a learning algorithm that is intended to reduce its generalization error but not its training error'' \cite{goodfellow2015explaining}. To that end, many regularization techniques have been developed that aim to improve the training process in neural networks. These techniques can be architectural in nature, such as Dropout \cite{dropout} and Batch Normalization \cite{batchnorm}, or they can alter some aspect of the training process, such as label smoothing \cite{inceptionv3} or the minimization of a weight norm \cite{hanson1989comparingbiases}. These techniques are briefly reviewed in this section, providing context for loss-function metalearning.

\subsection{Implicit biases in optimizers}

It may seem surprising that overparameterized neural networks are able to generalize at all, given that they have the capacity to memorize a training set perfectly, and in fact sometimes do (i.e., zero training error is reached). Different optimizers have different implicit biases that determine which solutions are ultimately found. These biases are helpful in providing implicit regularization to the optimization process \cite{neyshabur2015pathsgd}. Such implicit regularization is the result of a network norm---a measure of complexity---that is minimized as optimization progresses. This is why models continue to improve even after training set has been memorized (i.e., the training error global optima is reached) \cite{neyshabur2017geometry}.

For example, the process of stochastic gradient descent (SGD) itself has been found to provide regularization implicitly when learning on data with noisy labels \cite{blanc2020implicit}. In overparameterized networks, adaptive optimizers find very different solutions than basic SGD. These solutions tend to have worse generalization properties, even though they tend to have lower training errors \cite{wilson2017marginal}.

\subsection{Regularization approaches}

While optimizers may minimize a network norm implicitly, regularization approaches supplement this process and make it explicit. For example, a common way to restrict the parameter norm explicitly is through weight decay. This approach discourages network complexity by placing a cost on weights \cite{hanson1989comparingbiases}.

Generalization and regularization are often characterized at the end of training, i.e. as a behavior that results from the optimization process. Various findings have influenced work in regularization. For example, flat landscapes (that is, cases where perturbations to a model's parameters do not greatly affect loss) have better generalization properties \cite{keskar2016large,losslandscape,chaudhari2019entropysgd}. In overparameterized cases, the solutions at the center of these landscapes may have zero training error (i.e., perfect memorization), and under certain conditions, zero training error empirically leads to lower generalization error \cite{belkin2019reconciling, nakkiran2019deep}. However, when a training loss of zero is reached, generalization suffers \cite{ishida2020we}. This behavior can be thought of as overtraining, and techniques have been developed to reduce it at the end of the training process, such as early stopping \cite{morgan1990generalization} and flooding \cite{ishida2020we}.

Both early stopping and flooding assume that overfitting happens at the end of training, which is not always true \cite{golatkar2019time}. In fact, the order in which easy-to-generalize and hard-to-generalize concepts are learned is important for the network's ultimate generalization. For instance, larger learning rates early in the training process often lead to better generalization in the final model \cite{li2019biglr}. Similarly, low-error solutions found by SGD in a relatively quick manner---such as through high learning rates---often have good generalization properties \cite{yao2007early}.

Other techniques tackle overfitting by making it more difficult. Dropout \cite{dropout} makes some connections disappear. Cutout \cite{cutout}, Mixup \cite{mixup}, and their composition, CutMix \cite{cutmix}, augment training data with a broader variation of examples.

Thus, the term regularization refers to a diverse set of techniques intended to prevent the network from overfitting. At this point there is no general theory of how it can be done. Different techniques aim at this goal in different ways that often interact; for example, flooding invalidates performance gains from early stopping \cite{ishida2020we}. However, ultimately all regularization techniques that act upon the training process alter the gradients that result from the training loss. This observation suggests loss-function optimization might be an effective way to regularize the training process. As will be shown later, it may also serve as a starting point for developing a theory of regularization.



\subsection{General loss-function metalearning}

The idea of metalearning loss-functions has a long history, with many different approaches, and promising recent developments in practical settings.

First, in unsupervised representation learning, weight update rules for semi-supervised learning have been metalearned successfully \cite{metz2018meta}. The update rules were constrained to fit a biological neuron model and transferred well between tasks.

Second, in reinforcement learning, various actor-critic approaches have tackled learning a meta-critic neural network that can generate losses \cite{sung2017learning,zhou2020online}. Metalearned critic network techniques have also been applied outside of reinforcement learning to train better few-shot classifiers \cite{antoniou2019selfcritique}.

Third, prior work in evolutionary computation showed that metalearning various types of objectives is useful. For instance, in evolving policy gradients \cite{houthooft2018evolved}, the policy loss is not represented symbolically, but rather as a neural network that convolves over a temporal sequence of context vectors. In reward function search \cite{niekum2010genetic}, the task is framed as a genetic programming problem, leveraging PushGP \cite{push}. Other techniques include metalearning state-dependent loss functions for inverse dynamics models \cite{morse2020learning}, and using a trained network that is itself a metalearned loss function \cite{bechtle2019meta}. 

While successful, these approaches do not directly tackle the problem of optimizing loss functions for deep learning---a topic outlined in the next subsection.

\subsection{Loss-function metalearning for deep networks: GLO and TaylorGLO}
\label{sec:lossfn_metalearning}

Concrete loss-function metalearning for deep networks was first introduced by \cite{gonzalez2019glo} as an automatic way to find customized loss functions that optimize a performance metric for a model. Their technique, a genetic programming approach named GLO, demonstrated that learned loss functions are most powerful when they are customized to individual tasks and architectures. Different loss functions can take advantage of the different characteristics of each such setting.

While GLO was effectively able to evolve loss functions that outperform the cross-entropy loss, it has a relatively unconstrained search space and creates many functions that are not well behaved (e.g., they may have discontinuities). As a result, many candidates have to be discarded in a costly two-stage optimization process. Therefore, in subsequent work, GLO's search space was replaced by a multivariate Taylor polynomial \cite{taylorglo}. Loss functions created by this method, TaylorGLO, are thus more likely to be well-behaved. They also have a tunable complexity based on the order of the polynomials; third-order functions were identified to work best in practical settings. This fixed parameterization allows TaylorGLO to scale to models with millions of trainable parameters, including a variety of deep learning architectures in image classification tasks.

Indeed, TaylorGLO loss functions were shown empirically to improve generalization in such models \cite{taylorglo}. As expected, these functions often had the same Baikal shape as those discovered by GLO. An important further advantage of TaylorGLO, however, is that it lends itself to theoretical analysis. Thus, with TaylorGLO, it is possible to understand regularization in this special case. The resulting push-pull theory constitutes a general principle, and may be used in the future to develop a more general theory of regularization.  This theory is the main contribution of this paper.

The TaylorGLO parameterization and the evolutionary optimization approach that leverages it are discussed in Appendix~\ref{sec:taylorexpansions} and Section~\ref{sec:taylorglo_method}, respectively.

\section{Loss Functions as Multivariate Taylor expansions}
\label{sec:taylorexpansions}

\renewcommand{\Tx}[1][-\theta_0]{(x_i #1)}
\renewcommand{\Ty}[1][-\theta_1]{(y_i #1)}

The core of the TaylorGLO technique for evolving loss functions \cite{taylorglo} is a fixed-length parameterization based on multivariate Taylor expansions. Taylor expansions \cite{taylor1715methodus} are a well-known function approximator that can represent differentiable functions within the neighborhood of a point using a polynomial series. Below, the common univariate Taylor expansion formulation is presented, followed by a natural extension to arbitrarily-multivariate functions.

Given a $C^{k_{\text{max}}}$ smooth (i.e., first through $k_{\text{max}}$ derivatives are continuous), real-valued function, $f(x): \mathbb{R}\to\mathbb{R}$, a $k$th-order Taylor approximation at point $a\in \mathbb{R}$, $\hat{f}_k(x,a)$, where $0\leq k \leq k_{\text{max}}$, can be constructed as
\begin{equation}
\hat{f}_k(x,a) = \sum_{n=0}^k \frac{1}{n!} f^{(n)}(a) (x-a)^n    .
\end{equation}
Conventional, univariate Taylor expansions have a natural extension to arbitrarily high-dimensional inputs of $f$. Given a $C^{k_{\text{max}+1}}$ smooth, real-valued function, $f(\vec{x}): \mathbb{R}^n\to\mathbb{R}$, a $k$th-order Taylor approximation at point $\vec{a}\in \mathbb{R}^n$, $\hat{f}_k(\vec{x},\vec{a})$, where $0\leq k \leq k_{\text{max}}$, can be constructed. The stricter smoothness constraint compared to the univariate case allows for the application of Schwarz's theorem on equality of mixed partials, obviating the need to take the order of partial differentiation into account.

Let us define an $n$th-degree multi-index, $\alpha = (\alpha_1,\alpha_2,\ldots,\alpha_n)$, where $\alpha_i \in \mathbb{N}_0$, $|\alpha| = \sum_{i=1}^n \alpha_i$, $\alpha! = \prod_{i=1}^n \alpha_i!$. $\vec{x}^\alpha = \prod_{i=1}^n x_i^{\alpha_i}$, and $\vec{x} \in \mathbb{R}^n$. Multivariate partial derivatives can be concisely written using a multi-index
\begin{equation}
\partial^\alpha f = \partial_1^{\alpha_1}\partial_2^{\alpha_2}\cdots\partial_n^{\alpha_n} f = \frac{\partial^{|\alpha|}}{\partial x_1^{\alpha_1}\partial x_2^{\alpha_2}\cdots\partial x_n^{\alpha_n}}   .
\end{equation}
Thus, discounting the remainder term, the multivariate Taylor expansion for $f(\vec{x})$ at $\vec{a}$ is
\begin{equation}
\hat{f}_k(\vec{x},\vec{a}) = \sum_{\forall \alpha,|\alpha|\leq k} \frac{1}{\alpha!} \partial^{\alpha}f(\vec{a}) (\vec{x}-\vec{a})^\alpha    .
\label{eqn:taylorexpmultivariate}
\end{equation}
The unique partial derivatives in $\hat{f}_k$ and $\vec{a}$ are parameters for a $k$th order Taylor expansion. Thus, a $k$th order Taylor expansion of a function in $n$ variables requires $n$ parameters to define the center, $\vec{a}$, and one parameter for each unique multi-index $\alpha$, where $|\alpha|\leq k$. That is: 
\begin{equation}
\#_{\text{parameters}}(n,k) = n + {n+k\choose k} = n + \frac{(n+k)!}{n!\,k!}  \;.
\end{equation}


\label{sec:parameterization}

The multivariate Taylor expansion can be leveraged for a novel loss-function parameterization \cite{taylorglo}. 
Let an $n$-class classification loss function be defined as $\mathcal{L}_{\text{Log}} = -\frac{1}{n}\sum^{n}_{i=1} f(x_i,y_i)$.
The function $f(x_i,y_i)$ can be replaced by its $k$th-order, bivariate Taylor expansion, $\hat{f}_k(x,y,a_x,a_y)$. More sophisticated loss functions can be supported by having more input variables beyond $x_i$ and $y_i$, such as a time variable or unscaled logits (i.e., the raw outputs of a neural network's final fully-connected layer). This approach can be useful, for example, to evolve loss functions that change as training progresses.

For example, a loss function in $\vec{x}$ and $\vec{y}$ has the following third-order parameterization with parameters $\vec{\theta}$ (where $\vec{a} = \left< \theta_0, \theta_1 \right>$ and all other $\theta_x$ represent individual instances of $\partial^\alpha$ in Equation~\ref{eqn:taylorexpmultivariate}):
\begin{equation}
\label{eq:k3example}
\begin{aligned}
\mathcal{L}(\vec{x},\vec{y}) = -\frac{1}{n}\sum^n_{i=1} \Big[      \theta_2 + \theta_3\Ty + \tfrac{1}{2}\theta_4\Ty^2 \\+ \tfrac{1}{6}\theta_5\Ty^3 + \theta_6\Tx 
   + \theta_7\Tx\Ty \\+ \tfrac{1}{2}\theta_8\Tx\Ty^2 + \tfrac{1}{2}\theta_9\Tx^2 \\
   + \tfrac{1}{2}\theta_{10}\Tx^2\Ty + \tfrac{1}{6}\theta_{11}\Tx^3  \Big]
\end{aligned}
\end{equation}
Notably, the reciprocal-factorial coefficients can be integrated to be a part of the parameter set by direct multiplication if desired.

As was shown by \cite{taylorglo}, the technique makes it possible to train neural networks that are more accurate and learn faster than those with tree-based loss function representations. Representing loss functions in this manner confers several useful properties:
\begin{itemize}
\item It guarantees smooth functions;
\item Functions do not have poles (i.e., discontinuities going to infinity or negative infinity) within their relevant domain;
\item They can be implemented purely as compositions of addition and multiplication operations;
\item They can be trivially differentiated;
\item Nearby points in the search space yield similar results (i.e., the search space is locally smooth), making the fitness landscape easier to search;
\item Valid loss functions can be found in fewer generations and with higher frequency;
\item Loss function discovery is consistent and not dependent on a specific initial population; and
\item The search space has a tunable complexity parameter (i.e., the order of the expansion).
\end{itemize}

These properties are not necessarily held by alternative function approximators. For instance:
\begin{description}
\item[Fourier series] are well suited for approximating periodic functions \cite{fourier1829theorie}. Consequently, they are not as well suited for loss functions, whose local behavior within a narrow domain is important. Being a composition of waves, Fourier series tend to have many critical points within the domain of interest. Gradients fluctuate around such points, making gradient descent infeasible. Additionally, close approximations require a large number of terms, which in itself can be injurious, causing large, high-frequency fluctuations known as ``ringing'', due to Gibb's phenomenon \cite{wilbraham1848certain}.
\item[Pad\'{e} approximants] can be more accurate approximations than Taylor expansions; indeed, Taylor expansions are a special case of Pad\'{e} approximants where $M = 0$ \cite{graves1979numerical}. However, unfortunately Pad\'{e} approximants can model functions with one or more poles, which valid loss functions typically should not have. These problems still exist, and are exacerbated, for Chisholm approximants (a bivariate extension) \cite{chisholm1973rational}) and Canterbury approximants (a multivariate generalization) \cite{graves1975calculation}.
\item[Laurent polynomials] can represent functions with discontinuities, the simplest being $x^{-1}$. While Laurent polynomials provide a generalization of Taylor expansions into negative exponents, the extension is not useful because it results in the same issues as Pad\'{e} approximants.
\item[Polyharmonic splines] can represent continuous functions within a finite domain, however, the number of parameters is prohibitive in multivariate cases.
\end{description}

The multivariate Taylor expansion is therefore a better choice than the alternatives. It makes it possible to optimize loss functions efficiently in TaylorGLO.



\section{Learning rule decompositions for select loss functions}
\label{sec:learning_rule_decomp}
Substituting the {\bf Mean squared error (MSE)} loss into Equation~\ref{eqn:sgdsingleweight},
\begin{equation}
    \begin{aligned}
\hlamj \leftarrow \hlamj - \eta \frac{1}{n} \sum^n_{k=1}\bigg[  2 \left( h_k(\bm{x}_i,\hlam + s \bm{j}) - y_{ik} \right)  \dirdivhk  \bigg] \;\bigg\rvert_{s\rightarrow 0}
    \end{aligned}
\end{equation}
\begin{equation}
    \begin{aligned}
 =  \hlamj + \eta \frac{1}{n} \sum^n_{k=1}\bigg[  2 \left( y_{ik} - h_k(\bm{x}_i,\hlam) \right)  \dirdivhk \Big\rvert_{s\rightarrow 0}  \bigg],
    \end{aligned}
\end{equation}
and breaking up the coefficient expressions into $\gamma_k(\bm{x}_i,\bm{y}_i,\hlam)$ results in the weight update step
\begin{equation}
    \begin{aligned}
\gamma_k(\bm{x}_i,\bm{y}_i,\hlam) = 2y_{ik} - 2 h_k(\bm{x}_i,\hlam).
    \end{aligned}
\end{equation}
~\\
Substituting the {\bf Cross-entropy loss} into Equation~\ref{eqn:sgdsingleweight},
\begin{equation}
    \begin{aligned}
\hlamj \leftarrow \hlamj + \eta \frac{1}{n} \sum^n_{k=1}\bigg[  y_{ik} \frac{1}{h_k(\bm{x}_i,\hlam + s \bm{j})}  \dirdivhk  \bigg] \;\bigg\rvert_{s\rightarrow 0}
    \end{aligned}
\end{equation}
\begin{equation}
    \begin{aligned}
 =  \hlamj + \eta \frac{1}{n} \sum^n_{k=1}\bigg[   \frac{y_{ik} }{h_k(\bm{x}_i,\hlam)}  \dirdivhk \Big\rvert_{s\rightarrow 0}  \bigg],
    \end{aligned}
\end{equation}
and breaking up the coefficient expressions into $\gamma_k(\bm{x}_i,\bm{y}_i,\hlam)$ results in the weight update step
\begin{equation}
    \begin{aligned}
\gamma_k(\bm{x}_i,\bm{y}_i,\hlam) = \frac{y_{ik} }{h_k(\bm{x}_i,\hlam)}.
    \end{aligned}
\end{equation}
~\\
Substituting the {\bf Baikal loss} into Equation~\ref{eqn:sgdsingleweight},
\begin{equation}
    \begin{aligned}
\hlamj \leftarrow \hlamj + \eta \frac{1}{n} \sum^n_{k=1}\bigg[ \bigg( \frac{1}{h_k(\bm{x}_i,\hlam + s \bm{j})}    +  \frac{y_{ik}}{h_k(\bm{x}_i,\hlam + s \bm{j})^2}  \bigg) \dirdivhk  \bigg] \;\bigg\rvert_{s\rightarrow 0}
    \end{aligned}
\end{equation}
\begin{equation}
    \begin{aligned}
 =  \hlamj + \eta \frac{1}{n} \sum^n_{k=1}\bigg[   \left( \frac{1}{h_k(\bm{x}_i,\hlam)} +  \frac{y_{ik}}{h_k(\bm{x}_i,\hlam)^2}  \right)     \dirdivhk \Big\rvert_{s\rightarrow 0}  \bigg],
    \end{aligned}
\end{equation}
and breaking up the coefficient expressions into $\gamma_k(\bm{x}_i,\bm{y}_i,\hlam)$ results in the weight update step
\begin{equation}
    \begin{aligned}
\gamma_k(\bm{x}_i,\bm{y}_i,\hlam) = \frac{1}{h_k(\bm{x}_i,\hlam)} +  \frac{y_{ik}}{h_k(\bm{x}_i,\hlam)^2}.
    \end{aligned}
\end{equation}
~\\
Substituting the {\bf Third-order TaylorGLO loss} with parameters $\bm{\lambda}$ into Equation~\ref{eqn:sgdsingleweight},
\begin{equation}
    \begin{aligned}
\hlamj \leftarrow \hlamj + \eta \frac{1}{n} \sum^n_{k=1}\bigg[    \lambda_2 \dirdivhk  +  \lambda_3 2 \hkminuslamone \dirdivhk  \\[-0.6em]+  \lambda_4 3 \hkminuslamone^2 \dirdivhk  +  \lambda_5 (y_{ik} - \lambda_0) \dirdivhk  \\+ \left( \lambda_6 (y_{ik} - \lambda_0) 2 \hkminuslamone + \lambda_7  (y_{ik} - \lambda_0)^2 \right) \dirdivhk   \bigg] \;\bigg\rvert_{s\rightarrow 0}
    \end{aligned}
\end{equation}
\begin{equation}
    \begin{aligned}
 =  \hlamj + \eta \frac{1}{n} \sum^n_{k=1}\bigg[    \left(\lambda_3 + \lambda_6 (y_{ik} - \lambda_0)\right) 2 \hkminuslamone \dirdivhk \Big\rvert_{s\rightarrow 0}  \\[-0.6em]+  \left(\lambda_2 + \lambda_5 (y_{ik} - \lambda_0) + \lambda_7 (y_{ik} - \lambda_0)^2\right) \dirdivhk \Big\rvert_{s\rightarrow 0}  \\+  \lambda_4 3 \left(h_k(\bm{x}_i,\hlam) - \lambda_1 \right)^2 \dirdivhk \Big\rvert_{s\rightarrow 0}       \bigg],
    \end{aligned}
\end{equation}
and breaking up the coefficient expressions into $\gamma_k(\bm{x}_i,\bm{y}_i,\hlam)$ results in the weight update step
\begin{equation}
    \begin{aligned}
\gamma_k(\bm{x}_i,\bm{y}_i,\hlam) = \left(\lambda_3 + \lambda_6 (y_{ik} - \lambda_0)\right) 2 \left(h_k(\bm{x}_i,\hlam) - \lambda_1 \right)  \\+  \lambda_2 + \lambda_5 (y_{ik} - \lambda_0) + \lambda_7 (y_{ik} - \lambda_0)^2  +  \lambda_4 3 \left(h_k(\bm{x}_i,\hlam) - \lambda_1 \right)^2
    \end{aligned}
\end{equation}
\vspace{0.5em}
\begin{equation}
    \begin{aligned} 
 = 2 \lambda_3 h_k(\bm{x}_i,\hlam) - 2 \lambda_1 \lambda_3 + 2 \lambda_6 h_k(\bm{x}_i,\hlam) y_{ik} - 2 \lambda_6 \lambda_0 h_k(\bm{x}_i,\hlam) \\- 2\lambda_1 \lambda_6 y_{ik} + 2\lambda_1 \lambda_6 \lambda_0 + \lambda_2 + \lambda_5 y_{ik} - \lambda_5 \lambda_0 + \lambda_7 y_{ik}^2 - 2\lambda_7 \lambda_0 y_{ik} \\+ \lambda_7 \lambda_0^2 + 3 \lambda_4 h_k(\bm{x}_i,\hlam)^2 - 6 \lambda_1 \lambda_4 h_k(\bm{x}_i,\hlam) + 3 \lambda_4 \lambda_1^2 .
    \end{aligned}
\end{equation}
To simplify analysis in this case, $\gamma_k(\bm{x}_i,\bm{y}_i,\hlam)$ can be decomposed into a linear combination of
\begin{equation}
[1, h_k(\bm{x}_i,\hlam), h_k(\bm{x}_i,\hlam)^2, h_k(\bm{x}_i,\hlam)y_{ik}, y_{ik}, y_{ik}^2]
\end{equation}
with respective coefficients $[c_1, c_h, c_{hh}, c_{hy}, c_y, c_{yy} ]$ whose values are implicitly functions of $\bm{\lambda}$:
\begin{equation}
\label{eqn:tayk3_gamma_ccomb}
\gamma_k(\bm{x}_i,\bm{y}_i,\hlam) = c_1 + c_h h_k(\bm{x}_i,\hlam) + c_{hh} h_k(\bm{x}_i,\hlam)^2 + c_{hy} h_k(\bm{x}_i,\hlam) y_{ik} + c_y y_{ik} + c_{yy} y_{ik}^2.
\end{equation}
~\\
Using these decompositions, it is possible to characterize and compare training dynamics different loss functions, as is done in Section~\ref{sec:characterizingtrainingdynamics}.

\section{Baikal attractors}
\label{sec:baikalzeroerrnotattractor}

~\\
\emph{\textbf{Theorem~1.} Zero training error regions of the weight space are not attractors for the Baikal loss function.}

~\\
\emph{Proof}: Given that Baikal does tend to minimize training error to a large degree---otherwise it would be useless as a loss function since we are effectively assuming that the training data is in-distribution---we can observe what happens as we approach a point in parameter space that is arbitrarily-close to zero training error. Assume, without loss of generality, that all non-target scaled logits have the same value.
\begin{equation}
\hlamj \leftarrow \hlamj + \eta \frac{1}{n} \sum^n_{k=1} \left\{
        \begin{array}{rl}
            {\displaystyle \lim_{h_k(\bm{x}_i,\hlam)\to \frac{\epsilon}{n-1}}}\; \gamma_k(\bm{x}_i,\bm{y}_i,\hlam)  D_{\bm{j}} \left( h_k(\bm{x}_i,\hlam) \right) & \quad y_{ik} = 0 \\
            {\displaystyle \lim_{h_k(\bm{x}_i,\hlam)\to 1-\epsilon}}\; \gamma_k(\bm{x}_i,\bm{y}_i,\hlam)  D_{\bm{j}} \left( h_k(\bm{x}_i,\hlam) \right)  & \quad y_{ik} = 1
        \end{array}
    \right.
\end{equation}
\begin{equation}
= \hlamj + \eta \frac{1}{n} \sum^n_{k=1} \left\{
        \begin{array}{rl}
            {\displaystyle \lim_{h_k(\bm{x}_i,\hlam)\to \frac{\epsilon}{n-1}}}\;\;    \left(\dfrac{1}{h_k(\bm{x}_i,\hlam)} +  \dfrac{0}{h_k(\bm{x}_i,\hlam)^2}\right)      D_{\bm{j}} \left( h_k(\bm{x}_i,\hlam) \right) & \quad y_{ik} = 0 \\
            {\displaystyle \lim_{h_k(\bm{x}_i,\hlam)\to 1-\epsilon}}\;    \left(\dfrac{1}{h_k(\bm{x}_i,\hlam)} +  \dfrac{1}{h_k(\bm{x}_i,\hlam)^2}\right)      D_{\bm{j}} \left( h_k(\bm{x}_i,\hlam) \right)  & \quad y_{ik} = 1
        \end{array}
    \right.
\end{equation}
\begin{equation}
= \hlamj + \eta \frac{1}{n} \sum^n_{k=1} \left\{
        \renewcommand{\arraystretch}{2}
        \begin{array}{rl}
            \dfrac{n-1}{\epsilon}      D_{\bm{j}} \left( h_k(\bm{x}_i,\hlam) \right) & \quad y_{ik} = 0 \\
            \left(\dfrac{1}{1-\epsilon} +  \dfrac{1}{\left( 1-\epsilon \right)^2}\right)      D_{\bm{j}} \left( h_k(\bm{x}_i,\hlam) \right)  & \quad y_{ik} = 1
        \end{array}
    \right.
\end{equation}
\begin{equation}
= \hlamj + \eta \frac{1}{n} \sum^n_{k=1} \left\{
        \renewcommand{\arraystretch}{2}
        \begin{array}{rl}
            \dfrac{n-1}{\epsilon}      D_{\bm{j}} \left( h_k(\bm{x}_i,\hlam) \right) & \quad y_{ik} = 0 \\
            \dfrac{2 - \epsilon}{\epsilon^2 - 2\epsilon + 1}      D_{\bm{j}} \left( h_k(\bm{x}_i,\hlam) \right)  & \quad y_{ik} = 1
        \end{array}
    \right.
\end{equation}
The behavior in the $y_{ik} = 0$ case will dominate for small values of $\epsilon$. Both cases have a positive range for small values of $\epsilon$, ultimately resulting in non-target scaled logits becoming maximized, and subsequently the non-target logit becoming minimized. This is equivalent, in expectation, to saying that $\epsilon$ will become larger after applying the learning rule. A larger $\epsilon$ clearly implies a move away from a zero training error area of the parameter space. Thus, zero training error is not an attractor for the Baikal loss function. 
\qed

~\\
\section{Change in entropy}
\label{sec:softmaxentropyderivation}

~\\
\emph{\textbf{Theorem~2.}
The change in entropy is proportional to
\begin{equation}
	\dfrac{ \epsilon (\epsilon - 1) \left( \mathrm{e}^{\epsilon (\epsilon - 1) (\gamma_{\neg T} - \gamma_T)}      -       \mathrm{e}^{\frac{\epsilon (\epsilon - 1) \gamma_T (n-1) + \epsilon \gamma_{\neg T} (\epsilon (n-3) + n - 1)}{(n-1)^2}}               \right)       }{   (\epsilon - 1)\; \mathrm{e}^{\epsilon (\epsilon - 1) (\gamma_{\neg T} - \gamma_T)} - \epsilon\; \mathrm{e}^{\frac{\epsilon (\epsilon - 1) \gamma_T (n-1) + \epsilon \gamma_{\neg T} (\epsilon (n-3) + n - 1)}{(n-1)^2}}          } ,
\end{equation}
where $\gamma_{\neg T}$ is the value of $\gamma_j$ for non-target logits, and $\gamma_T$ for the target logit.}

~\\
\emph{Proof}: Let us analyze the case where all non-target logits have the same value, $\frac{\epsilon}{n-1}$, and the target logit has the value $1-\epsilon$. That is, all non-target classes have equal probabilities.

A model's scaled logit for an input $\bm{x}_i$ can be represented as:
\begin{equation}
h_k(\bm{x}_i,\hlam) = \sigma_k(f(\bm{x}_i,\hlam)) = \frac{\mathrm{e}^{f_k(\bm{x}_i,\hlam)}}{\sum_{j=1}^{n} \mathrm{e}^{f_j(\bm{x}_i,\hlam)}}
\end{equation}
where $f_k(\bm{x}_i,\hlam)$ is a raw output logit from the model.

The $(k,j)$th entry of the Jacobian matrix for $h(\bm{x}_i,\hlam)$ can be easily derived through application of the chain rule:
\begin{equation}
\bm{J}_{kj} h(\bm{x}_i,\hlam) = \frac{\partial h_k(\bm{x}_i,\hlam)}{\partial f_j(\bm{x}_i,\hlam)} =\left\{
        \begin{array}{rl}
           h_j(\bm{x}_i,\hlam)\; (1 - h_k(\bm{x}_i,\hlam))\;f_k(\bm{x}_i,\hlam) & \quad k=j \\
           - h_j(\bm{x}_i,\hlam) \;h_k(\bm{x}_i,\hlam)\;f_k(\bm{x}_i,\hlam) & \quad k\neq j
        \end{array}
    \right.
\end{equation}
Consider an SGD learning rule of the form:
\begin{equation}
\hlamj \leftarrow \hlamj + \eta \frac{1}{n} \sum^n_{k=1}\left[ \gamma_k(\bm{x}_i,\bm{y}_i,\hlam) D_{\bm{j}} \left( h_k(\bm{x}_i,\hlam) \right)   \right]
\end{equation}
Let us freeze a network at any specific point during the training process for any specific sample. Now, treating all $f_j(\bm{x}_i,\hlam), j\in [1,n]$ as free parameters with unit derivatives, rather than as functions. That is, $\theta_j = f_j(\bm{x}_i,\hlam)$. We observe that updates are as follows:
\begin{equation}
\Delta f_j \propto \sum^n_{k=1} \gamma_j \left\{
        \begin{array}{rl}
           h_j(\bm{x}_i,\hlam)\; (1 - h_k(\bm{x}_i,\hlam)) & \quad k=j \\
           - h_j(\bm{x}_i,\hlam) \;h_k(\bm{x}_i,\hlam) & \quad k\neq j
        \end{array}
    \right.
\end{equation}
For downstream analysis, we can consider, as substitutions for $\gamma_j$ above, $\gamma_{\neg T}$ to be the value for non-target logits, and $\gamma_T$ for the target logit.

This sum can be expanded and conceptually simplified by considering $j$ indices and $\neg j$ indices. $\neg j$ indices, of which there are $n-1$, are either all non-target logits, or one is the target logit in the case where $j$ is not the target logit. Let us consider both cases, while substituting the scaled logit values defined above:
\begin{equation}
\Delta f_j \propto  \left\{
        \begin{array}{rl}
           \gamma_{\neg T} \;\bm{J}_{k=j} h(\bm{x}_i,\hlam)      + (n-2) \gamma_{\neg T} \;\bm{J}_{k\neq j} h(\bm{x}_i,\hlam)   + \gamma_{T} \;\bm{J}_{k\neq j} h(\bm{x}_i,\hlam)            & \quad \text{non-target}\;j \\
           \gamma_T \;\bm{J}_{k=j} h(\bm{x}_i,\hlam)  + (n-1) \gamma_{\neg T} \;\bm{J}_{k\neq j} h(\bm{x}_i,\hlam)            & \quad  \text{target}\;j
        \end{array}
    \right.
\end{equation}
\begin{equation}
\Delta f_j \propto  \left\{
        \begin{array}{rl}
           \gamma_{\neg T} h_{\neg T}(\bm{x}_i,\hlam)\; (1 - h_{\neg T}(\bm{x}_i,\hlam))      \\+ (n-2) \gamma_{\neg T} \left(- h_{\neg T}(\bm{x}_i,\hlam) \;h_{\neg T}(\bm{x}_i,\hlam)\right)   \\+ \gamma_{T} \left(- h_{\neg T}(\bm{x}_i,\hlam) \;h_T(\bm{x}_i,\hlam)\right)            & \quad \text{non-target}\;j \\\\
           \gamma_T h_T(\bm{x}_i,\hlam)\; (1 - h_T(\bm{x}_i,\hlam))  \\+ (n-1) \gamma_{\neg T} \left(- h_{\neg T}(\bm{x}_i,\hlam) \;h_T(\bm{x}_i,\hlam)\right)           & \quad  \text{target}\;j
        \end{array}
    \right.
\end{equation}
\begin{equation}
\text{where}\quad  h_T(\bm{x}_i,\hlam) = 1-\epsilon ,  \quad   h_{\neg T}(\bm{x}_i,\hlam) = \frac{\epsilon}{n-1}
\end{equation}
\begin{equation}
\Delta f_j \propto  \left\{
        \begin{array}{rl}
           \gamma_{\neg T} \dfrac{\epsilon}{n-1} \bigg(1-\frac{\epsilon}{n-1}\bigg)  +  \gamma_{\neg T} (n-2)\dfrac{\epsilon^2}{n^2-2n+1} + \gamma_T (\epsilon-1)\dfrac{\epsilon}{n-1}         & \quad \text{non-target}\;j \\
           \gamma_T \epsilon - \gamma_T \epsilon^2 + \gamma_{\neg T}(n-1)(\epsilon-1) \dfrac{\epsilon}{n-1}         & \quad  \text{target}\;j
        \end{array}
    \right.
\end{equation}
At this point, we have closed-form solutions for the changes to softmax inputs. To characterize entropy, we must now derive solutions for the changes to softmax outputs given such changes to the inputs. That is:
\begin{equation}
\Delta \sigma_j(f(\bm{x}_i,\hlam)) = \frac{\mathrm{e}^{f_j(\bm{x}_i,\hlam) + \Delta f_j}}{\sum_{k=1}^{n} \mathrm{e}^{f_k(\bm{x}_i,\hlam)+\Delta f_k}}
\end{equation}
Due to the two cases in $\Delta f_j$, $\Delta \sigma_j(f(\bm{x}_i,\hlam))$ is thus also split into two cases for target and non-target logits:
\begin{equation}
\Delta \sigma_j(f(\bm{x}_i,\hlam)) = \left\{
        \renewcommand{\arraystretch}{2}
        \begin{array}{rl}
           \dfrac{\mathrm{e}^{f_{\neg T}(\bm{x}_i,\hlam) + \Delta f_{\neg T}}}{(n-1) \mathrm{e}^{f_{\neg T}(\bm{x}_i,\hlam) + \Delta f_{\neg T}} + \mathrm{e}^{f_T(\bm{x}_i,\hlam) + \Delta f_T}}         & \quad \text{non-target}\;j \\
           \dfrac{\mathrm{e}^{f_{T}(\bm{x}_i,\hlam) + \Delta f_T}}{(n-1) \mathrm{e}^{f_{\neg T}(\bm{x}_i,\hlam) + \Delta f_{\neg T}} + \mathrm{e}^{f_T(\bm{x}_i,\hlam) + \Delta f_T}}         & \quad  \text{target}\;j
        \end{array}
    \right.
\end{equation}
Now, we can see that scaled logits have a lower entropy distribution when $\Delta \sigma_T(f(\bm{x}_i,\hlam)) > 0$ and $\Delta \sigma_{\neg T}(f(\bm{x}_i,\hlam)) < 0$. Essentially, the target and non-target scaled logits are being repelled from each other. We can ignore either of these inequalities, if one is satisfied then both are satisfied, in part because $|\bm{\sigma}(f(\bm{x}_i,\hlam))|_1 = 1$. The target-case constraint (i.e., the target scaled logit must grow) can be represented as:
\begin{equation}
	\label{eqn:taytargetconstraintforzeroerrattractor}
	\dfrac{\mathrm{e}^{f_{T}(\bm{x}_i,\hlam) + \Delta f_T}}{(n-1) \mathrm{e}^{f_{\neg T}(\bm{x}_i,\hlam) + \Delta f_{\neg T}} + \mathrm{e}^{f_T(\bm{x}_i,\hlam) + \Delta f_T}} > 1-\epsilon
\end{equation}
Consider the target logit case prior to changes:
\begin{equation}
	\dfrac{\mathrm{e}^{f_{T}(\bm{x}_i,\hlam)}}{(n-1) \mathrm{e}^{f_{\neg T}(\bm{x}_i,\hlam)} + \mathrm{e}^{f_T(\bm{x}_i,\hlam)}} = 1-\epsilon
\end{equation}
Let us solve for $\mathrm{e}^{f_{T}(\bm{x}_i,\hlam)}$:
\begin{eqnarray}
\mathrm{e}^{f_{T}(\bm{x}_i,\hlam)} &=& (n - 1) \mathrm{e}^{f_{\neg T}(\bm{x}_i,\hlam)} + \mathrm{e}^{f_{T}(\bm{x}_i,\hlam)} - \epsilon (n - 1) \mathrm{e}^{f_{\neg T}(\bm{x}_i,\hlam)} - \epsilon \mathrm{e}^{f_{T}(\bm{x}_i,\hlam)} \\
&=& \left( \frac{n-1}{\epsilon} - n + 1 \right) \mathrm{e}^{f_{\neg T}(\bm{x}_i,\hlam)}
\end{eqnarray}
Substituting this definition into Equation~\ref{eqn:taytargetconstraintforzeroerrattractor}:
\begin{equation}
	\dfrac{     \mathrm{e}^{\Delta f_T} \left( \dfrac{n-1}{\epsilon} - n + 1 \right) \mathrm{e}^{f_{\neg T}(\bm{x}_i,\hlam)}          }{(n-1) \mathrm{e}^{f_{\neg T}(\bm{x}_i,\hlam) + \Delta f_{\neg T}} +           \mathrm{e}^{\Delta f_T} \left( \dfrac{n-1}{\epsilon} - n + 1 \right) \mathrm{e}^{f_{\neg T}(\bm{x}_i,\hlam)}        } > 1-\epsilon
\end{equation}
Coalescing exponents:
\begin{equation}
	\dfrac{     \mathrm{e}^{\Delta f_T + f_{\neg T}(\bm{x}_i,\hlam)} \left( \dfrac{n-1}{\epsilon} - n + 1 \right)          }{(n-1) \mathrm{e}^{f_{\neg T}(\bm{x}_i,\hlam) + \Delta f_{\neg T}} +           \mathrm{e}^{\Delta f_T + f_{\neg T}(\bm{x}_i,\hlam)} \left( \dfrac{n-1}{\epsilon} - n + 1 \right)         } + \epsilon - 1 > 0
\end{equation}
Substituting in definitions for $\Delta f_T$ and $\Delta f_{\neg T}$ and greatly simplifying in a CAS is able to remove instances of $f_{\neg T}$:
\begin{equation}
	\dfrac{ \epsilon (\epsilon - 1) \left( \mathrm{e}^{\epsilon (\epsilon - 1) (\gamma_{\neg T} - \gamma_T)}      -       \mathrm{e}^{\dfrac{\epsilon (\epsilon - 1) \gamma_T (n-1) + \epsilon \gamma_{\neg T} (\epsilon (n-3) + n - 1)}{(n-1)^2}}               \right)       }{   (\epsilon - 1) \mathrm{e}^{\epsilon (\epsilon - 1) (\gamma_{\neg T} - \gamma_T)} - \epsilon \mathrm{e}^{\dfrac{\epsilon (\epsilon - 1) \gamma_T (n-1) + \epsilon \gamma_{\neg T} (\epsilon (n-3) + n - 1)}{(n-1)^2}}            } > 0
\end{equation}
\qed

~\\
\section{Implicit label smoothing}
\label{sec:labelsmoothinggeneraltaylor}

~\\
\emph{\textbf{Theorem~3.}
For any $\bm{\lambda}$ and any $\alpha \in (0,1)$, there exists a $\bm{\hat{\lambda}}$ such that the behavior imposed by $\bm{\hat{\lambda}}$ without explicit label smoothing is identical to the behavior imposed by $\bm{\lambda}$ \emph{with} explicit label smoothing.
}

~\\
\newcommand{\lsynontarget}{\dfrac{\alpha}{n} }
\newcommand{\lsynontargetsqr}{\dfrac{\alpha^2}{n^2} }
\newcommand{\lsytarget}{\left(1-\alpha\dfrac{n-1}{n}\right) }
\emph{Proof}: Consider a basic setup with standard label smoothing, controlled by a hyperparameter $\alpha \in (0,1)$, such that the target value in any $\bm{y}_i$ is $1-\alpha\frac{n-1}{n}$, rather than $1$, and non-target values are $\frac{\alpha}{n}$, rather than $0$. The learning rule changes in the general case as follows:
\begin{equation}
    \begin{aligned}
\gamma_k(\bm{x}_i,\bm{y}_i,\hlam) = \left\{
	\renewcommand{\arraystretch}{2}
        \begin{array}{rl}
	        c_1 + c_h h_k(\bm{x}_i,\hlam) + c_{hh} h_k(\bm{x}_i,\hlam)^2 \\+ c_{hy} h_k(\bm{x}_i,\hlam) \lsynontarget + c_y \lsynontarget + c_{yy} \lsynontargetsqr & \;\; y_{ik} = 0 \\\\
	        c_1 + c_h h_k(\bm{x}_i,\hlam) + c_{hh} h_k(\bm{x}_i,\hlam)^2 + c_{hy} h_k(\bm{x}_i,\hlam) \lsytarget \\+ c_y \lsytarget + c_{yy} \lsytarget^2   & \;\; y_{ik} = 1
        \end{array}
    \right.
    \end{aligned}
\end{equation}
Let $\hat{c}_1, \hat{c}_h, \hat{c}_{hh}, \hat{c}_{hy}, \hat{c}_y, \hat{c}_{yy}$ represent settings for $c_1, c_h, c_{hh}, c_{hy}, c_y, c_{yy}$ in the non-label-smoothed case that implicitly apply label smoothing within the TaylorGLO parameterization. Given the two cases in the label-smoothed and non-label-smoothed definitions of $\gamma_k(\bm{x}_i,\bm{y}_i,\hlam)$, there are two equations that must be satisfiable by settings of $\hat{c}$ constants for any $c$ constants, with shared terms highlighted in blue and red:
\begin{equation}
	\label{eqn:lstaygeneralzerocase}
	\begin{aligned}
{\color{blue} c_1 + c_h h_k(\bm{x}_i,\hlam) + c_{hh} h_k(\bm{x}_i,\hlam)^2 }+ c_{hy} h_k(\bm{x}_i,\hlam) \lsynontarget + c_y \lsynontarget + c_{yy} \lsynontargetsqr    \\=    {\color{red} \hat{c}_1 + \hat{c}_h h_k(\bm{x}_i,\hlam) + \hat{c}_{hh} h_k(\bm{x}_i,\hlam)^2 }
	\end{aligned}
\end{equation}
\begin{equation}
    \label{eqn:lstaygeneralonecase}
	\begin{aligned}
{\color{blue} c_1 + c_h h_k(\bm{x}_i,\hlam) + c_{hh} h_k(\bm{x}_i,\hlam)^2 }+ c_{hy} h_k(\bm{x}_i,\hlam) \lsytarget \\+ c_y \lsytarget + c_{yy} \lsytarget^2    \\=    {\color{red} \hat{c}_1 + \hat{c}_h h_k(\bm{x}_i,\hlam) + \hat{c}_{hh} h_k(\bm{x}_i,\hlam)^2 }+ \hat{c}_{hy} h_k(\bm{x}_i,\hlam) + \hat{c}_y + \hat{c}_{yy}
	\end{aligned}
\end{equation}
Let us then factor the left-hand side of Equation~\ref{eqn:lstaygeneralzerocase} in terms of different powers of $h_k(\bm{x}_i,\hlam)$:
\begin{equation}
    \label{eqn:lstaygeneralc1chchhdefs}
	\begin{aligned}
\underbrace{\left(c_1 + c_y \lsynontarget + c_{yy} \lsynontargetsqr\right)}_{\hat{c}_1} + \underbrace{\left(c_h + c_{hy} \lsynontarget\right)}_{\hat{c}_h} h_k(\bm{x}_i,\hlam) + \underbrace{c_{hh}}_{\hat{c}_{hh}} h_k(\bm{x}_i,\hlam)^2
	\end{aligned}
\end{equation}
Resulting in definitions for $\hat{c}_1, \hat{c}_h, \hat{c}_{hh}$. Let us then add the following form of zero to the left-hand side of Equation~\ref{eqn:lstaygeneralonecase}:
\begin{equation}
	\begin{aligned}
\left( c_{hy} h_k(\bm{x}_i,\hlam) \lsynontarget + c_y \lsynontarget + c_{yy} \lsynontargetsqr \right) - \left( c_{hy} h_k(\bm{x}_i,\hlam) \lsynontarget + c_y \lsynontarget + c_{yy} \lsynontargetsqr \right)
	\end{aligned}
\end{equation}
This allows us to substitute the definitions for $\hat{c}_1, \hat{c}_h, \hat{c}_{hh}$ from Equation~\ref{eqn:lstaygeneralc1chchhdefs} into Equation~\ref{eqn:lstaygeneralonecase}:
\begin{equation}
	\begin{aligned}
{\color{red}\hat{c}_1 + \hat{c}_h h_k(\bm{x}_i,\hlam) + \hat{c}_{hh} h_k(\bm{x}_i,\hlam)^2}    - \left( c_{hy} h_k(\bm{x}_i,\hlam) \lsynontarget + c_y \lsynontarget + c_{yy} \lsynontargetsqr \right)       \\+      c_{hy} h_k(\bm{x}_i,\hlam) \lsytarget + c_y \lsytarget + c_{yy} \lsytarget^2    \\=    {\color{red} \hat{c}_1 + \hat{c}_h h_k(\bm{x}_i,\hlam) + \hat{c}_{hh} h_k(\bm{x}_i,\hlam)^2 }+ \hat{c}_{hy} h_k(\bm{x}_i,\hlam) + \hat{c}_y + \hat{c}_{yy}
	\end{aligned}
\end{equation}
Simplifying into:
\begin{equation}
    \label{eqn:lstaygeneralalmostthereonecase}
	\begin{aligned}
c_{hy} h_k(\bm{x}_i,\hlam) \lsytarget + c_y \lsytarget + c_{yy} \lsytarget^2    \\- \left( c_{hy} h_k(\bm{x}_i,\hlam) \lsynontarget + c_y \lsynontarget + c_{yy} \lsynontargetsqr \right)   \\=    \hat{c}_{hy} h_k(\bm{x}_i,\hlam) + \hat{c}_y + \hat{c}_{yy}
	\end{aligned}
\end{equation}
Finally, factor the left-hand side of Equation~\ref{eqn:lstaygeneralalmostthereonecase} in terms of, $h_k(\bm{x}_i,\hlam)$, $1$, and $1^2$:
\begin{equation}
	\begin{aligned}
\underbrace{\left(   c_{hy} \lsytarget - c_{hy} \lsynontarget  \right)}_{\hat{c}_{hy}} h_k(\bm{x}_i,\hlam) \\+ \underbrace{\left(    c_{y} \lsytarget - c_{y} \lsynontarget     \right)}_{\hat{c}_{y}} + \underbrace{\left(     c_{yy} \lsytarget^2 - c_{yy} \lsynontargetsqr    \right)}_{\hat{c}_{yy}}
	\end{aligned}
\end{equation}


Thus, the in-parameterization constants with implicit label smoothing can be defined for any desired, label-smoothed constants as follows:
\begin{eqnarray}
\hat{c}_{1} &=& c_1 + c_y \lsynontarget + c_{yy} \lsynontargetsqr \\
\hat{c}_{h} &=& c_h + c_{hy} \lsynontarget \\
\hat{c}_{hh} &=& c_{hh} \\
\hat{c}_{hy} &=& c_{hy} \lsytarget - c_{hy} \lsynontarget \\
\hat{c}_{y} &=& c_{y} \lsytarget - c_{y} \lsynontarget \\
\hat{c}_{yy} &=& c_{yy} \lsytarget^2 - c_{yy} \lsynontargetsqr
\end{eqnarray}
So for any $\bm{\lambda}$ and any $\alpha \in (0,1)$, there exists a $\bm{\hat{\lambda}}$ such that the behavior imposed by $\bm{\hat{\lambda}}$ without explicit label smoothing is identical to the behavior imposed by $\bm{\lambda}$ \emph{with} explicit label smoothing. That is, any degree of label smoothing can be implicitly represented for any TaylorGLO loss function.
\qed

~\\
\section{Trainability of TaylorGLO loss functions}
\label{sec:tayinvariantderivation}

~\\
\emph{\textbf{Theorem~4.}
A third-order TaylorGLO loss function is not trainable if the following constraints on $\bm{\lambda}$ are satisfied:}
\begin{eqnarray}
c_1 + c_y + c_{yy} + \dfrac{c_h+c_{hy}}{n} + \dfrac{c_{hh}}{n^2}  &<& \left(n-1\right)\left(c_1 + \dfrac{c_h}{n} + \dfrac{c_{hh}}{n^2} \right) \\
c_y + c_{yy} + \dfrac{c_{hy}}{n}  &<& \left(n-2\right)\left(c_1 + \dfrac{c_h}{n} + \dfrac{c_{hh}}{n^2} \right).
\end{eqnarray}
~\\
\emph{Proof}: At the null epoch, a valid loss function aims to, in expectation, minimize non-target scaled logits while maximizing target scaled logits. Thus, we attempt to find cases of $\bm{\lambda}$ for which these behaviors occur. Considering the representation for $\gamma_k(\bm{x}_i,\bm{y}_i,\hlam)$ in Equation~\ref{eqn:tayk3_gamma_ccomb}:
\begin{equation}
\hlamj \leftarrow \hlamj + \eta \frac{1}{n} \sum^n_{k=1} \left\{
		\renewcommand{\arraystretch}{1.3}
        \begin{array}{rl}
            \left(   c_1 + c_h h_k(\bm{x}_i,\hlam) + c_{hh} h_k(\bm{x}_i,\hlam)^2   \right) D_{\bm{j}} \left( h_k(\bm{x}_i,\hlam) \right) & \quad y_{ik} = 0 \\
            \big(    c_1 + c_h h_k(\bm{x}_i,\hlam) + c_{hh} h_k(\bm{x}_i,\hlam)^2 \\+ c_{hy} h_k(\bm{x}_i,\hlam) + c_y + c_{yy}   \big) D_{\bm{j}} \left( h_k(\bm{x}_i,\hlam) \right)  & \quad y_{ik} = 1.
        \end{array}
    \right.
\end{equation}
Let us substitute $h_k(\bm{x}_i,\hlam) = \frac{1}{n}$ (i.e., the expected value of a logit at the null epoch):
\begin{equation}
\hlamj \leftarrow \hlamj + \eta \frac{1}{n} \sum^n_{k=1} \left\{
		\renewcommand{\arraystretch}{2}
        \begin{array}{rl}
            \left(   c_1 + \dfrac{c_h}{n} + \dfrac{c_{hh}}{n^2}   \right) D_{\bm{j}} \left( h_k(\bm{x}_i,\hlam) \right) & \quad y_{ik} = 0 \\
            \left(    c_1 + c_y + c_{yy} + \dfrac{c_h+c_{hy}}{n} + \dfrac{c_{hh}}{n^2}   \right) D_{\bm{j}} \left( h_k(\bm{x}_i,\hlam) \right)  & \quad y_{ik} = 1.
        \end{array}
    \right.
\end{equation}
For the desired degenerate behavior to appear, the directional derivative's coefficient in the $y_{ik} = 1$ case must be less than zero:
\begin{eqnarray}
c_1 + c_y + c_{yy} + \frac{c_h + c_{hy}}{n} + \frac{c_{hh}}{n^2}  &<& 0.
\end{eqnarray}
This finding can be made more general, by asserting that the directional derivative's coefficient in the $y_{ik} = 1$ case be less than $(n-1)$ times the coefficient in the $y_{ik} = 0$ case. Thus, for a loss function to be viable it has to satisfy the following constraint on $\bm{\lambda}$:
\begin{eqnarray}
c_1 + c_y + c_{yy} + \dfrac{c_h+c_{hy}}{n} + \dfrac{c_{hh}}{n^2}  &<& \left(n-1\right)\left(c_1 + \dfrac{c_h}{n} + \dfrac{c_{hh}}{n^2} \right) \\
c_y + c_{yy} + \dfrac{c_{hy}}{n}  &<& \left(n-2\right)\left(c_1 + \dfrac{c_h}{n} + \dfrac{c_{hh}}{n^2} \right).
\end{eqnarray}
\qed

\section{Experimental setup and code}
\label{sec:experimental_setup_appendix}

A full implementation of TaylorGLO and experiment configuration files are available at \url{https://github.com/ANONYMIZED/ANONYMIZED}. 
Configuration files are available in the ``experiments'' directory. Those experiments that are run with the invariant on TaylorGLO parameters (as described in Section~\ref{sec:invariant_section}) are identified by ``Invar'' in their name. The details on the repository's structure and commands through which the experiments can be run are specified in the repository's README file. The specific experiments in this paper are described below (for further details, see the README file).

\subsection{Domains}
 The CIFAR-10 \cite{krizhevsky2009learning} image classification dataset was used to illustrate the results in this paper. Note that the original TaylorGLO paper also validated the technique against the MNIST \cite{mnist}, CIFAR-100 \cite{krizhevsky2009learning}, and SVHN \cite{svhn} datasets.

\subsection{Evaluated architectures}
This paper includes results on four different types of deep neural network architectures: AlexNet \cite{NIPS2012_4824}, AllCNN-C \cite{allcnn}, Preactivation ResNet-20 \cite{preresnet}, which is an improved variant of the ubiquitous ResNet architecture \cite{resnet}, and Wide ResNets of different morphologies \cite{wideresnet}. Results from an experiment with Cutout \cite{cutout} regularization were also included, to reinforce that TaylorGLO provides a different, complementary approach to regularization.

Models were trained with the same hyperparameters as specified in the literature. Inputs were normalized by subtracting their mean pixel value and dividing by their pixel standard deviation. Standard data augmentation techniques consisting of random horizontal flips and croppings with two pixel padding were applied during training.


\subsection{TaylorGLO setup}
CMA-ES was instantiated with population size $\lambda=40$ for experiments with the derived constraint and $\lambda=20$ without it. An initial step size $\sigma=1.2$ was used throughout. These values are the same as those in the original TaylorGLO paper.

\subsection{Implementation details}

Due to the number of partial training sessions that are
needed to evaluate TaylorGLO loss function candidates, training
was distributed across the network to a cluster, composed of dedicated machines with NVIDIA GeForce GTX 1080Ti GPUs. Training itself was implemented with
PyTorch \cite{pytorch} in Python. The primary components of TaylorGLO
(i.e., the genetic algorithm and CMA-ES) were implemented in the
Swift programming language, which allows for easy parallelization. These components run centrally on one machine and dispatch work asynchronously to the cluster. 

Following the original TaylorGLO paper, training for each candidate was aborted and retried up to two additional times if validation accuracy was below 0.15 at the tenth epoch. This method helped reduce computation costs.

\end{document}